\documentclass[18pt]{report}
\usepackage[english]{babel}

\usepackage{fullpage}
\usepackage{titlesec}  
\usepackage{titling}
\usepackage{algcompatible}
\titleformat{\chapter}[display]
{\normalfont%
   \Large 
    \bfseries}{\chaptertitlename\ \thechapter}{18pt}{%
    }
\usepackage{amsmath,amssymb}

\usepackage{amsmath,amssymb}
\usepackage[hypertexnames=false]{hyperref}
\usepackage{wrapfig}
\usepackage{mwe}
\usepackage{lipsum}
\usepackage{acronym}
\usepackage{subfigure}
\numberwithin{equation}{section}
\usepackage{psfrag}
\usepackage{float}
\usepackage{times}
\usepackage[intoc]{nomencl}
\usepackage{physics}
\usepackage[utf8]{inputenc}
\usepackage{multicol}
\usepackage{psfrag} 
\usepackage{amsmath}

\usepackage[draft,enable-survey]{pdfpages}
\usepackage{pdfpages}
\usepackage{graphicx}
\usepackage{geometry}
\usepackage[style=numeric, sorting=none]{biblatex}
\usepackage[thinc]{esdiff}
\usepackage{float}
\usepackage[utf8]{inputenc} 
\usepackage[T1]{fontenc}    
\usepackage{amssymb,amsmath,amsfonts} 
\usepackage{braket} 
\usepackage{caption}
\usepackage{algorithm}
\usepackage[noend]{algpseudocode}
\usepackage{gensymb}
\usepackage{libertine}
\usepackage{wrapfig}
\usepackage[algo2e]{algorithm2e}
\usepackage{booktabs}

\begin{document}
\begin{titlepage}
\pagenumbering{empty}
\begin{center}
\Large\textsc{\textbf{Design and Development of Autonomous Delivery Robot}}\\
\textsc{} \\

\large{{Aniket Gujarathi, Akshay Kulkarni, Unmesh Patil, Yogesh Phalak, Rajeshree Deotalu, Aman Jain,  Navid Panchi}}

\textsc{} \\
\large\text{under the guidance of} \\
\textsc{} \\
\large{{Dr. Ashwin Dhabale} \\ and \\ {Dr. Shital S. Chiddarwar}}\\
\textsc{} \\
\begin{figure}[b]
\centering
\includegraphics[scale=1.5]{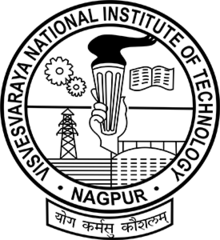}
\end{figure}
\vspace{4in}
\Large{\textbf{Visvesvaraya National Institute of Technology}}\\
\Large{\textbf{Nagpur 440 010(India)}}\\
\Large{\textbf{2020}}\\
\small{\textcopyright} \small{Visvesvaraya National Institute of Technology (VNIT) 2020}

\textsc{} \\
\end{center}
\end{titlepage}
\chapter*{ABSTRACT}

\par  The field of autonomous robotics is growing at a rapid rate. The trend to use increasingly more sensors in vehicles is driven both by legislation and consumer demands for higher safety and reliable service. Nowadays, robots are found everywhere, ranging from homes, hospitals to industries, and military operations. Autonomous robots are developed to be robust enough to work beside humans and to carry out jobs efficiently. Humans have a natural sense of understanding of the physical forces acting around them like gravity, sense of motion, etc. which are not taught explicitly but are developed naturally. However, this is not the case with robots. To make the robot fully autonomous and competent to work with humans, the robot must be able to perceive the situation and devise a plan for smooth operation, considering all the adversities that may occur while carrying out the tasks. In this thesis, we present an autonomous mobile robot platform that delivers the package within the VNIT campus without any human intercommunication. From an initial user-supplied geographic target location, the system plans an optimized path and autonomously navigates through it. The entire pipeline of an autonomous robot working in outdoor environments is explained in detail in this thesis. We have addressed the problem of semantic segmentation for road and obstacle detection. The common networks used in the literature are reported, along with some motivation for each of them. A general requirement for autonomous navigation is the availability of a high-definition map. The different layers of maps are discussed and a map of the VNIT campus with required details is constructed. The issue of robust localization and sensor fusion is explained in detail in this thesis. The problem of the need of 360-degree vision to the autonomous vehicles is also discussed. Catadioptric cameras that output panoramic views images with very large fields of view. It turns out that the design of such cameras solves plenty of problems including creating a 3D point cloud and providing preliminary visual data to the obstacle detection and motion planning. The proposed solution is characterized by an intricate mixture of optics and geometry as exemplified.

\pagenumbering{roman}

\addcontentsline{toc}{chapter}{Abstract}

\listoffigures
\addcontentsline{toc}{chapter}{List of Figures}

\listoftables
\addcontentsline{toc}{chapter}{List of Tables}

\tableofcontents 
\pagenumbering{gobble}

\cleardoublepage

\pagenumbering{arabic}

\chapter{Introduction}

The field of autonomous robots is growing rapidly in the world, in terms of both the diversity of emerging applications and the levels of interest among traditional players in the automotive, truck, public transportation, industrial, and military communities. Autonomous robotic systems offer the potential for significant enhancements in safety and operational efficiency. Due to the meteoric growth of e-commerce, developing faster, more affordable and sustainable last-mile deliveries become more important. Many challenges like reduced capacity, driver shortage, damaged and stolen products, failed delivery attempts, increased traffic congestion, etc. can be solved using autonomous robots. An autonomous robot is designed and engineered to deal with its environment on its own, and work for extended periods of time without human intervention. It must not only carry out its task of delivery properly, but must also consider the various scenarios changing around it and act accordingly. The robot must make quick decisions even in adverse conditions, considering the safety of pedestrians around it\cite{safety}. The aim of autonomous robots is to work alongside humans and try to make human life easier. 

\par Currently, many robots are being used in industries \cite{industry}, homes \cite{home}, military applications, disaster management \cite{apna_paper2}, etc., all around the world. The advancements in robotics has made lives easier for humans in many aspects and it provides with a safer and more efficient alternative to perform tasks which are difficult or time consuming for humans. Some of the applications of autonomous robots include cleaning robots like Roomba, delivery robots, autonomous vehicles, and other robots that move freely around a physical space without being guided by humans \cite{apna_paper1}.

\par
\begin{figure}[h]
\includegraphics[width=2.5in]{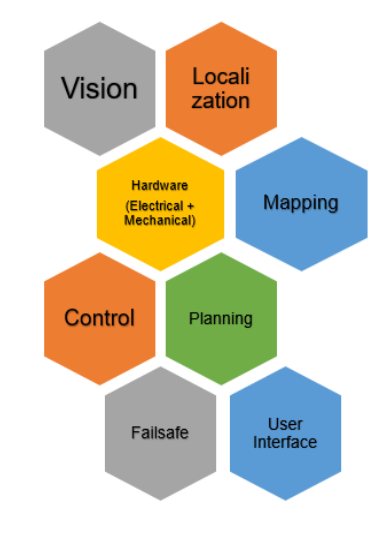}
\centering
\caption{System Pipeline}
\label{fig:archi}
\end{figure}

 In order to make a robot completely autonomous, the robot must be completely cognizant of its surroundings and must be able to perform actions based on the inputs it receives through various modules of the system. For the purpose of achieving a state of complete autonomy, the robot must be able to take information from sensors, perceive the environment, localize itself precisely in the world, and finally devise an optimal plan to achieve its goal. These instructions achieved from the modules mentioned above must be integrated by the robot in real-time and be given to a control node to actually move the system in the real world. The system pipeline of an autonomous robot is shown in Fig. \ref{fig:archi}.

The accuracy and the proper integration of all the modules is of utmost importance for an autonomous robot to operate. A fault in any of the module may cause serious repercussions and may even pose a hazard to humans around it. This thesis aims to implement all the mentioned modules flawlessly in order to achieve a completely autonomous operation of the robot in outdoor environments. 
\par This thesis is organized as follows :
\par In Chapter \ref{chapter:hardware}, the hardware design criteria with the applicable constraints is depicted. Furthermore, the hardware structure including the cyberphysical architecture of the robot is described. The schematic of the power system as well as the renderings of CAD models are illustrated.

\par Maps are an integral part of an autonomous operation pipeline. Delivery robots need an accurate map suitable for accurate localization, navigation and planning. Furthermore, these maps have to be able to incorporate and respond to the changes in the environment. The standard maps like Google Maps, used by humans to navigate the world cannot be used for the purpose of navigation of an autonomous robot. Chapter \ref{chapter:mapping} illustrates the construction of specialized maps for the purpose of autonomous navigation.

\par Chapter \ref{chapter:localization} describes the problem of state estimation and localization of a robot in detail. In order to navigate accurately around the world, the robot must know its location in the world and the map exactly. A robot can move smoothly only if it is properly localized. An inaccurate localization may cause the robot to vary off the roads or behave erroneously which are serious issues when the robot is completely autonomous.

\par The planning module is the backbone of an autonomous driving pipeline. A planner is responsible for finding optimal paths in the map for the robot to move and generating efficient trajectories and velocity profile for the robot to move locally in the presence of static/dynamic obstacles, obey lane rules, prioritize safety of humans, etc. Chapter \ref{chapter:planning} describes the hierarchical planning structure adopted for the autonomous driving pipeline and the implementation details of mission planner and local planner.

Chapter \ref{chapter:seg} addresses the problem of semantic segmentation for road and obstacle detection. This problem involves separating sets of pixels from an image where each separate set has some common attributes. Due to the complexity of the task and the availability of large datasets (with images and corresponding labels), most modern techniques use Supervised Deep Learning. Thus, some of the common networks used in the literature are described, along with some motivation for each of them. Following this, implementation and results of road segmentation are given.

Chapter \ref{chapter:catadioptric} addresses first in great depth the problem of the need of 360-degree vision to the autonomous vehicles, designing new solutions including catadioptric cameras that output panoramic views of the scene, i.e., images with very large fields of view. It turns out that the design of such cameras solves plenty of problems including creating a 3D point cloud and providing preliminary visual data to the obstacle detection and motion planning. The proposed solution is characterized by an intricate mixture of optics and geometry as exemplified in the second and third sections of the Chapter \ref{chapter:catadioptric}. 

Chapter \ref{chapter:camera_lidar} describes the process of intrinsic and extrinsic calibration of the camera-Lidar system.
\chapter{Hardware Design}
\label{chapter:hardware}
\section{Overview}

\par The robot is designed to operate as an autonomous mobile robot platform for different applications like in industrial areas for transportation, in hospitals for carrying food and medicines, in unmanned missions and also for security purposes. Our aim is to develop an autonomous delivery mobile robot that can deliver a package autonomously from A to B within the VNIT campus. In this chapter, the design criteria and system architecture are described.

\section{Design Criteria}

\par There were five major constraints in hardware design. These are outlined below:

\begin{enumerate}
    \item Modularity: In order to easily add units or parts to the robot, the robot platform has to be modular. These units or parts may be additional navigation sensors, room for payload, extra on-board power, or various devices for effective human-machine interface. 
    
    \item Low-cost Production: Even though mobile robots are available in the market, they tend to be expensive, thus increasing research and development costs. Further, the use of a ready-made robot will increase the cost of production even more in case of mass volume production.
    
    \item Truncated Construction: In the prototyping phase, the construction is kept simple and truncated in order to use minimal resources and to focus on designated functionality.
    
    \item Suitability of Environmental Conditions: Robot is planned to be used in outdoor environments, which means that the robot has to move on the roads and be able to pass over small obstacles. Additionally, electronic equipment on the robot must be protected.
    
    \item Originality: To contribute to scientific research and development, the robot has to be different and new. 
\end{enumerate}
The hardware and software structure of the robot has been designed by taking into consideration the above criteria.

\section{Hardware Structure}

\subsection{Hardware Overview}

\par
\begin{figure}[h]
\includegraphics[width=0.9\textwidth]{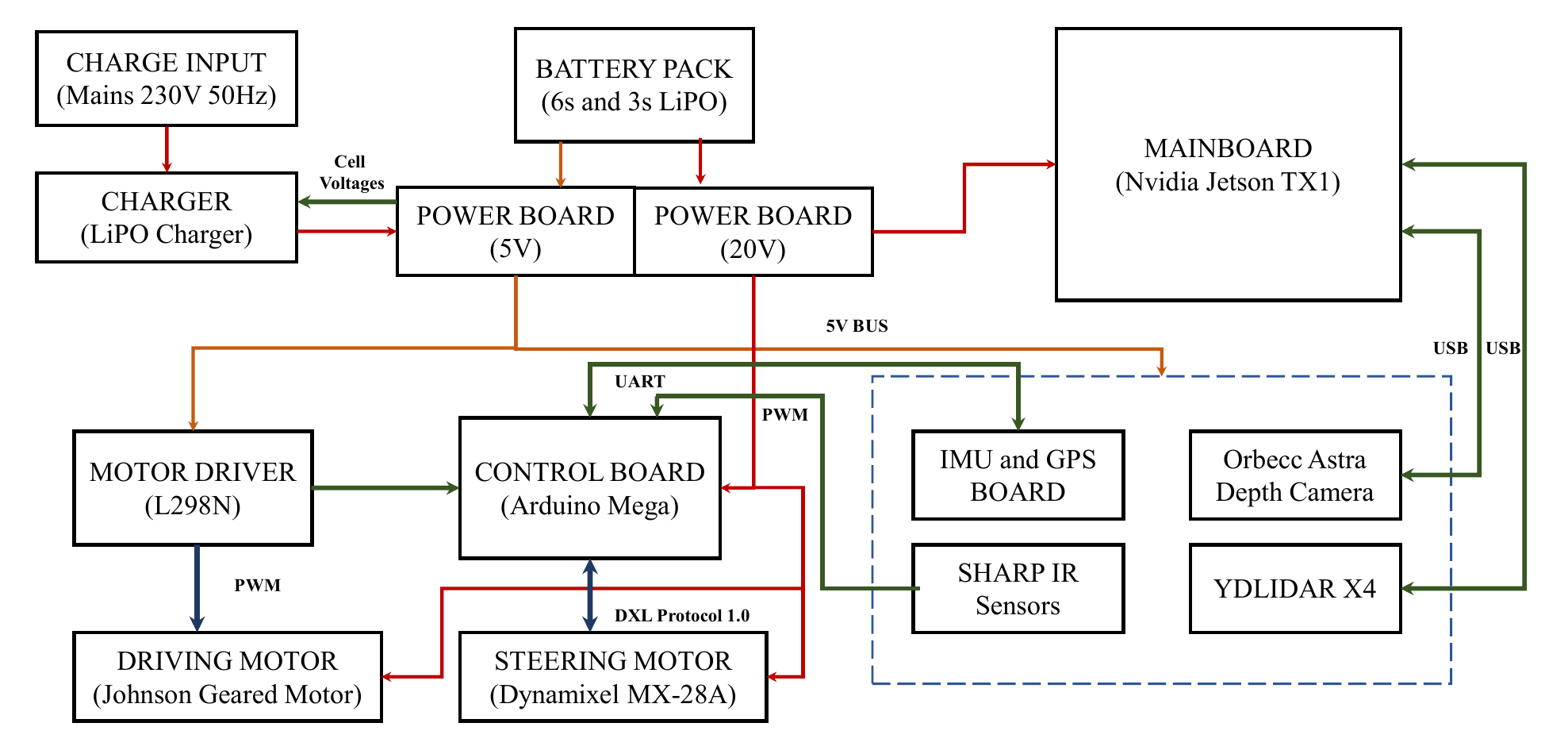}
\centering
\caption[Cyberphysical Architecture]{Cyberphysical Architecture of the Robot}
\label{fig:hard_Struct}
\end{figure}

 The hardware structure of the robot was made by considering the design requirements. Fig. \ref{fig:hard_Struct} shows the anatomy of the robot. Green lines symbolize signal and communication connections, while red and orange lines are main power connections and blue lines are connections between motors and motor drivers. In this section, every part of the robot is described below according to the design progress.

\subsection{Driving System}

Firstly, the driving system was designed for the robot. The driving structure is made up of the chassis of the robot, geared motor, Dynamixel smart servo motor, and motor driver. Pre-built chassis was used to accelerate the design and implementation steps. Chassis consists of an aluminum alloy skeleton, spring suspension, gearbox, and Ackerman steering mechanism. It was feasible to turn the available driving shaft by adding a gear to the chassis' gearbox. The 300 rpm Johnson geared motor is placed with the 1:1 gear ratio on the custom made motor mount to drive the shaft. Similarly, the steering mechanism is operated by the MX-28 Dynamixel Smart servo motor. 

\begin{figure}[h]
\includegraphics[width=\textwidth]{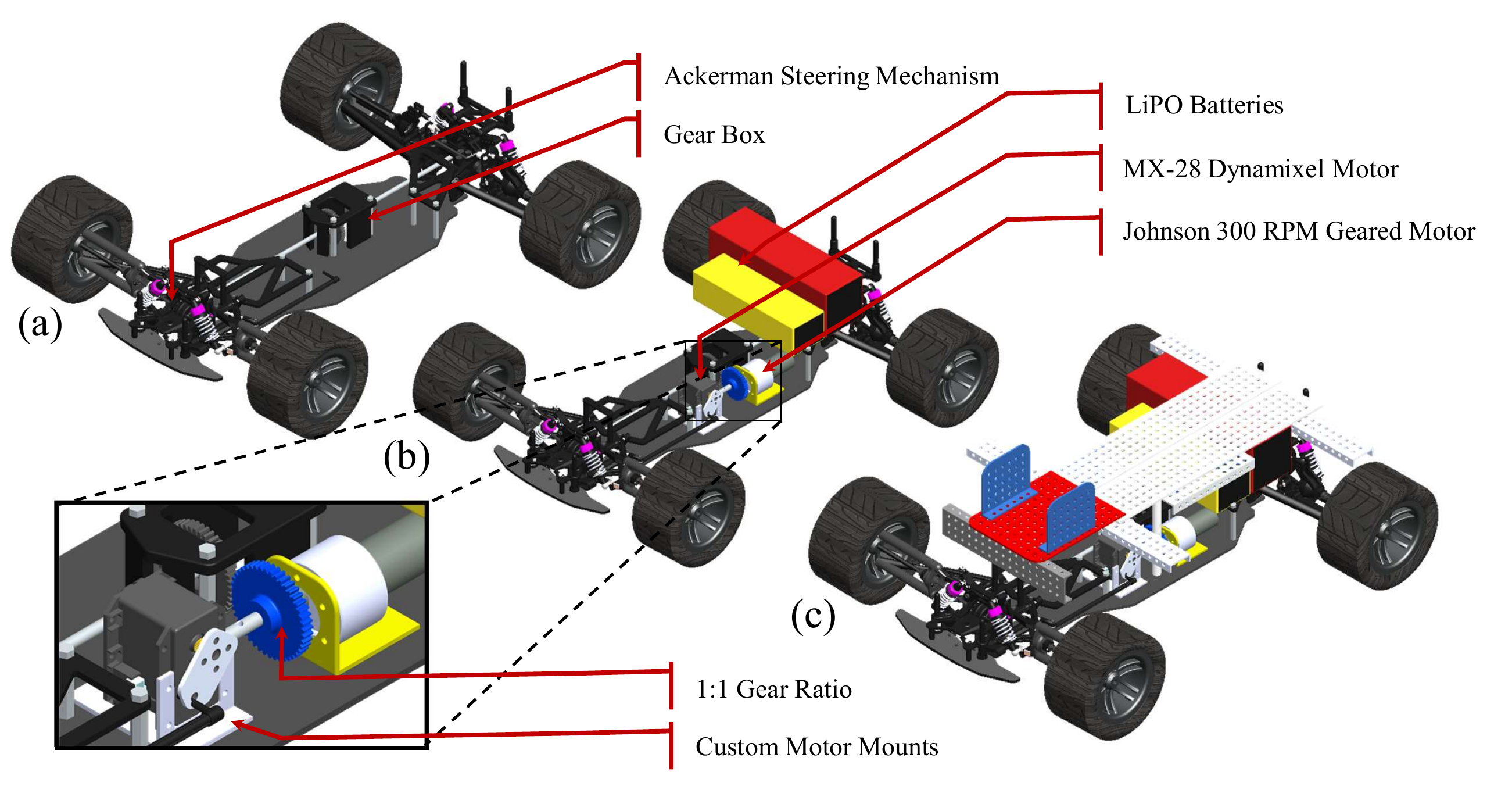}
\centering
\caption[Pre-built chassis of 1:10 scaled RC car]{(a) Pre-built chassis of 1:10 scaled RC car, (b) Mounted motors and LiPO batteries, (c) Aluminium plate covering frame for stage 1. }
\label{fig:chassis}
\includegraphics[width=\textwidth]{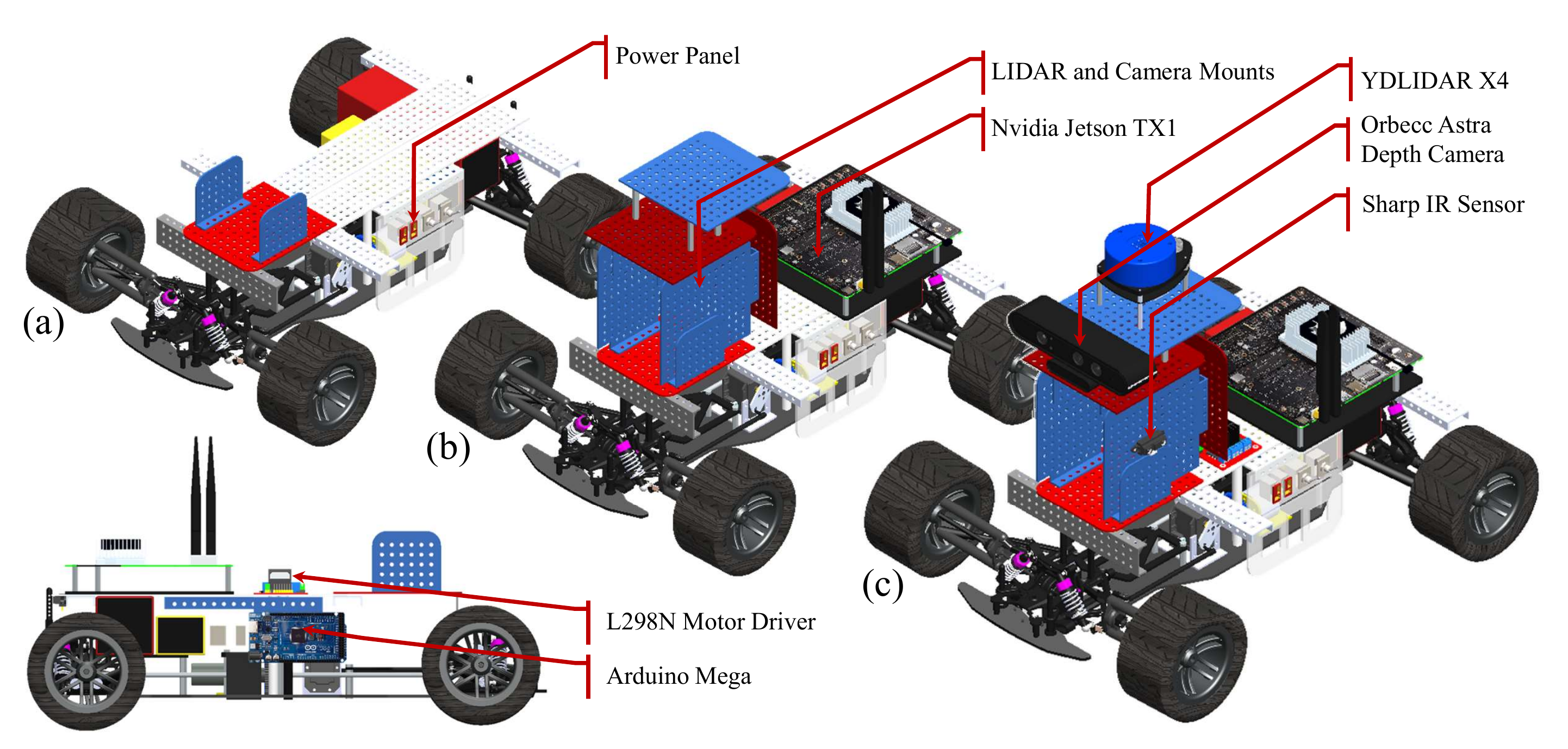}
\centering
\caption[Final Version of the Robot]{(a) Power panel mounted on the frame, (b) LIDAR and Camera mounts, (c) Final version of the Robot }
\label{fig:chassis1}
\end{figure}

\subsection{Power System}

\begin{figure}[h]
\includegraphics[width=\textwidth]{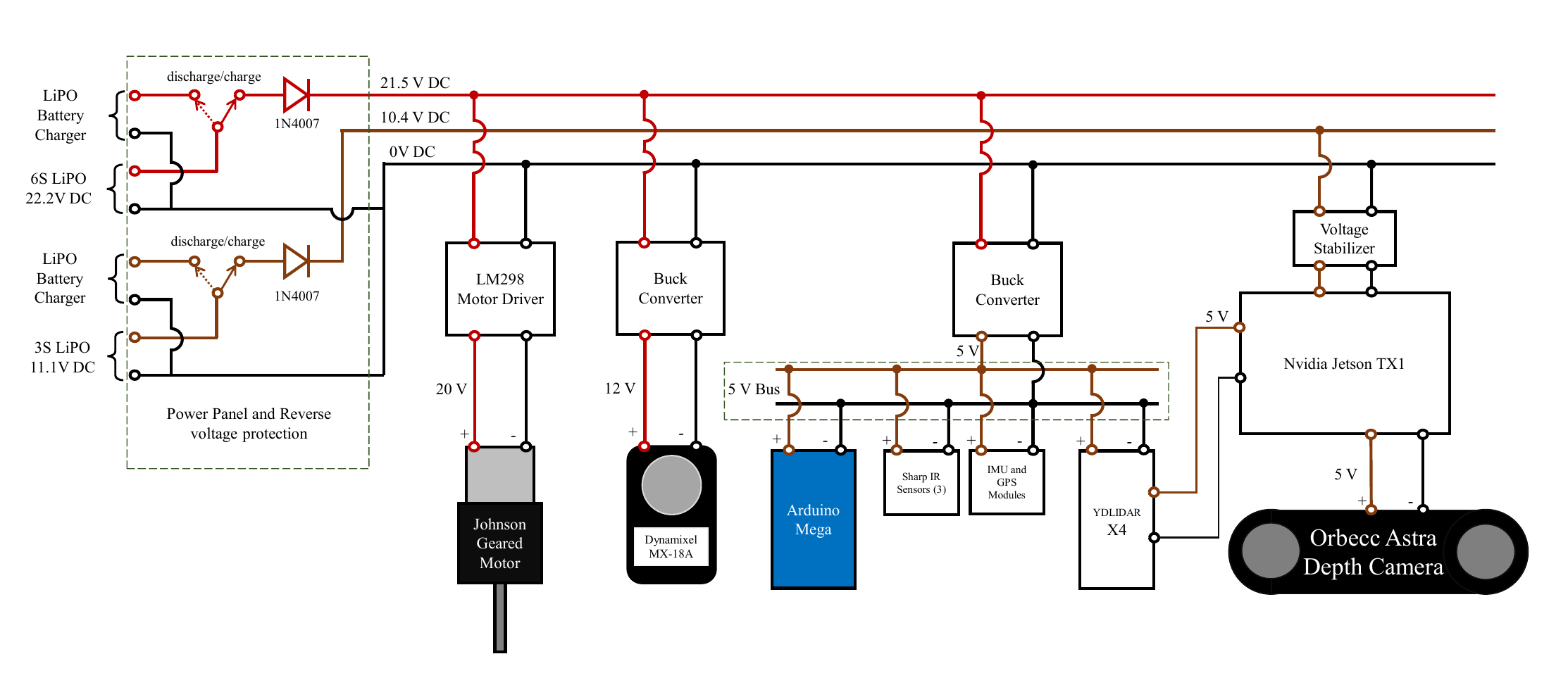}
\centering
\caption[Power diagram of the robot]{Power diagram of the robot}
\label{fig:power}
\end{figure}

The power system was designed according to the requirements of the drive system. The Power system has three parts. These parts are battery pack, power board, and charger. It has two Lithium Polymer (LiPo) battery packs to power its system (6s and 3s). LiPo batteries have higher capacity compared to other battery types in the same size and weight. But LiPo batteries have safety issues. For this reason, LiPo batteries have to be monitored while charging and discharging. The 5V bus is added to the system to power the auxiliaries such as sensors and motor drivers. The battery management system monitors battery status by measuring battery voltages, battery temperatures, and the current which is drawn from the battery pack. The power board MCU in LiPo charger cuts the power in the event of a dangerous situation, such as overvoltage or short-circuit, while charging or discharging batteries. The power panel is provided on the side of the robot to place ON/OFF switches and shifting between charging/discharging modes. The power system schematic of the robot is as given in the Fig. \ref{fig:power}.

\subsection{Controller and driver boards}

\par The total three controllers and driver boards mounted on the robot are given as follows.

\begin{enumerate}
    \item Mainboard (Nvidia Jetson TX1):  Mainboard is the brain of the robot. Nvidia Jetson TX1 is used for higher-level control. ROS is used for communication between the various modules. It's Nvidia Maxwell GPU (256 CUDA cores) enables fast inference times for deep neural networks. It runs all other processes like the global planning algorithm, the local planning algorithm, control algorithm, and so on. Further stages of the work, control methods and sensor fusion algorithms will be implemented on the mainboard.
    
    \item Control board (Arduino Mega):  Arduino Mega is used for lower-level control. It controls both the motors (main drive and steering), and handles all the sensors. It communicates with the main processor (Nvidia Jetson TX1) through rosserial to get commands for the motors and to publish sensor data.
    
    \item DC motor driver (L298): The L298 is an integrated monolithic circuit in a 15-lead Multiwatt and PowerSO20 packages. It is a high voltage, high current dual full-bridge driver designed to accept standard TTL logic levels and drive inductive loads such as relays, solenoids, DC and stepping motors. Two enable inputs are provided to enable or disable the device independently of the input signals. The emitters of the lower transistors of each bridge are connected and the corresponding external terminal can be used for the connection of an external sensing resistor. An additional supply input is provided so that the logic works at a lower voltage. The Johnsons geared DC motor is driven by this board.

\end{enumerate}

\subsection{Sensors}

\begin{figure}[h]
\includegraphics[width=0.8\textwidth]{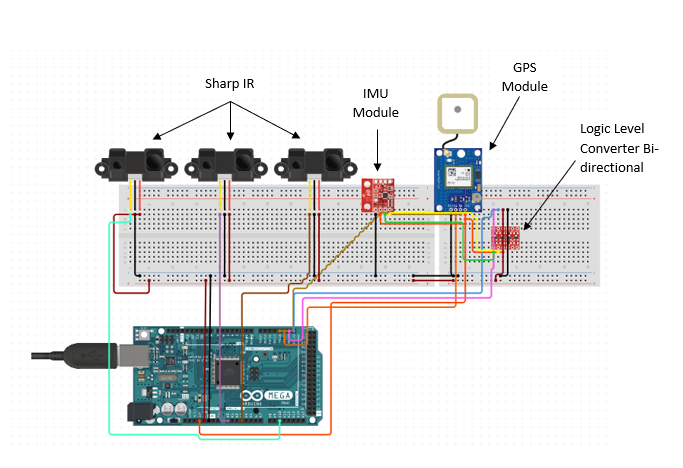}
\centering
\caption[Circuit diagram of the sensors.]{Connection schematic circuit diagram of the sensors.}
\label{fig:sensor}
\end{figure}

\par The sensors used in the robot are as follows: 

\begin{enumerate}
    \item Orbecc Astra Depth Camera: RGBD camera mounted to get front view image and front 3D depth map for perception and planning. It has a range of 8 meters for the 3D depth map. Useful for localization (using visual odometry) and for identifying obstacles.
    
    \item YDLIDAR X4: Laser range finder which gives a 2D (planar) 360 degrees depth map, used for perception and planning. It has a 10 meters scanning range. Useful for localization of robot and identifying obstacles.
    
    \item SparkFun IMU Breakout MPU9250: An inertial measurement unit (IMU). It consists of a 3-axis accelerometer, 3-axis gyroscope, and a 3-axis magnetometer. Useful for localization of robot.
    
    \item SHARP IR Sensor: A distance measuring sensor, to be used as the last line of defense against collisions. It has a range of 4-30 cm.
    
    \item Neo-M8N GPS Module: Gives global position (latitude, longitude, and altitude) useful for global planning. The used Neo-M8N GPS Module provides 167 dBm navigation sensitivity and supports all satellite augmentation systems.

\end{enumerate}

The connection schematic circuit diagram of the non optical sensors (except depth camera and LiDAR) is given in Fig. \ref{fig:sensor}.

\subsection{Level Design}

\par The hardware design of the robot is performed at three levels. As seen from Fig. \ref{fig:hard_Struct}, there are too many parts on the robot, and one of the design criteria is modularity. These levels are as follows:

\begin{enumerate}
   
    \item Body level (Fig. \ref{fig:chassis} b): This level consists of robot chassis, motors, batteries, temperature sensors, and control board. Parts in the body level are stationary and unique to the robot platform.

    \item Control unit level (Fig. \ref{fig:chassis1} b): Powerboard, mainboard, 5V DC bus, motor driver board, and distance sensors are placed at this level. This level is detachable, so that changes can be made. Control unit level can be used on any other platform, as long as the motor driver and the battery are fitted.

    \item Rooftop level (Fig. \ref{fig:chassis1} c): This level is designed for optical sensors. The camera and LiDAR is placed in this level to ensure no blockage in the range. In later stages of this work, additional navigation components and application-specific equipment can be added to this level.
    
\end{enumerate}

\section{Conclusion}

\begin{figure}[h]
\includegraphics[width=\textwidth]{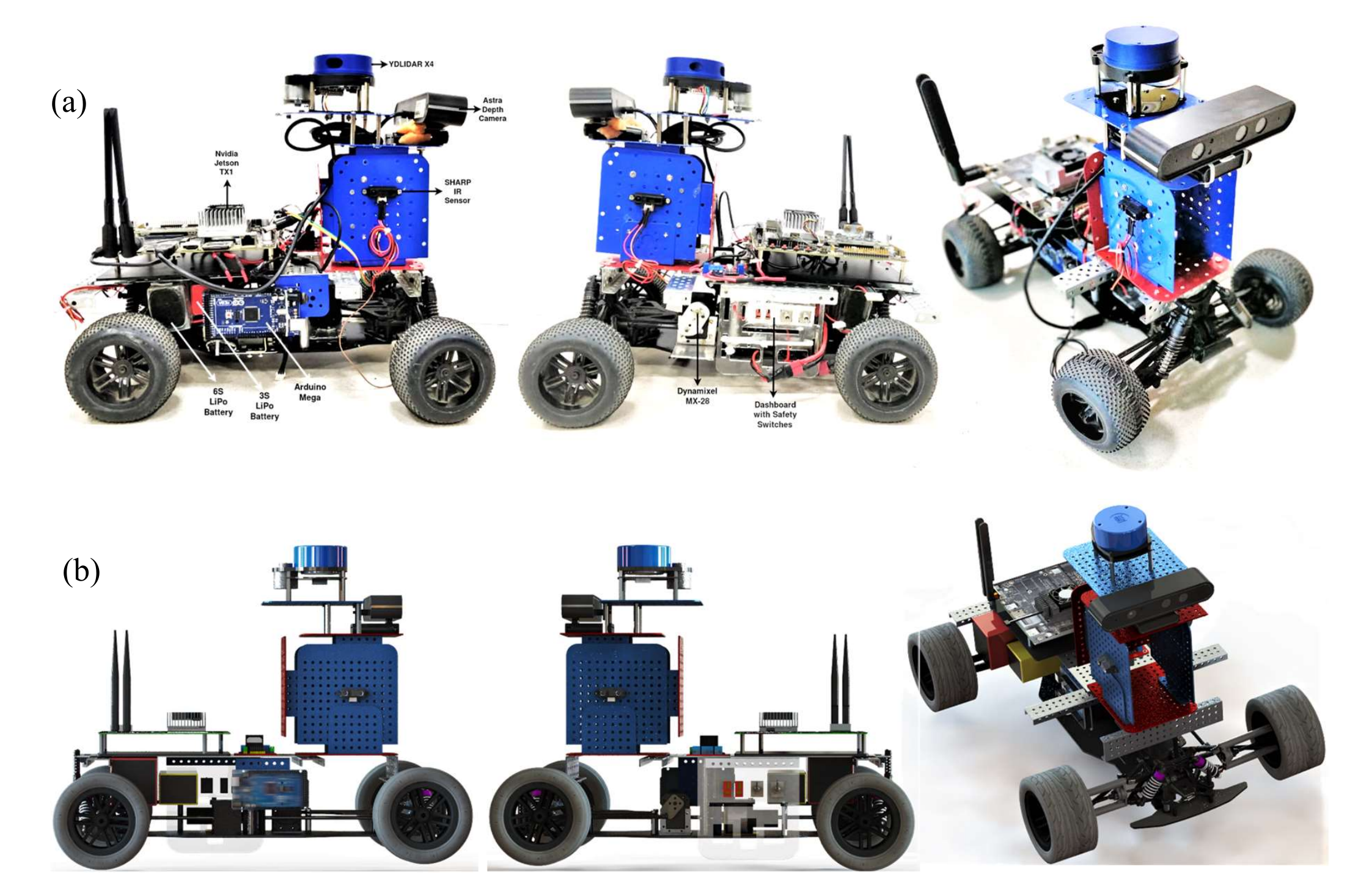}
\centering
\caption[The Autonomous delivery robot.]{(a) The final hardware of the Autonomous delivery robot, (b) The rendered images of respective CAD models.}
\label{fig:finalHardware}
\end{figure}

In conclusion, the robot which is seen in Fig. \ref{fig:finalHardware} is designed and built according to design criteria and open field tests are started. The environmental considerations are fully met and a robust structure has been developed. The robot weighs about 5kg and has a payload capacity of 2kg.
\chapter{Building and Managing Maps for Autonomous Operation}
\label{chapter:mapping}
\section{Overview}
\label{sec:mapping_overview}
One of the major aspects for the navigation of autonomous robots in outdoor environments is the availability of a specialized high-definition (HD) maps. Many web map services like Google Maps in existence are designed specifically for the purpose of humans to navigate the world. However, such maps offer a location resolution of up to a few meters which cannot be used for the purpose of navigating an autonomous robot due to safety reasons, errors, lack of details and information, etc. HD-maps also known as ADAS maps or Vector maps are generally used for autonomous navigation applications. Some benefits of HD-maps are :
\begin{enumerate}
 \item High accuracy of object locations, upto 10cm.
 \item Consist of multiple layers of information about the lanes, which way they travel, road intersections, curbs, 3D point cloud of the environment, etc.
\end{enumerate}
Using the information in the HD-maps, the robot can localize itself in the map and plan a path for navigating around the environment. However, existence of a map prior to starting of the operation is not always necessary. Another way of solving this problem of localization and mapping is through SLAM (Simultaneous Localization and Mapping) \cite{slam}. The purpose of SLAM is to generate a map and using the information in the map to simultaneously deduce the location of the robot in the map. The SLAM problem is still an active field of research and is widely used in autonomous robots ranging from indoor robots to outdoor robots, airborne systems and underwater robots. 
\par
\begin{figure}[t]
\includegraphics[width=4.5in]{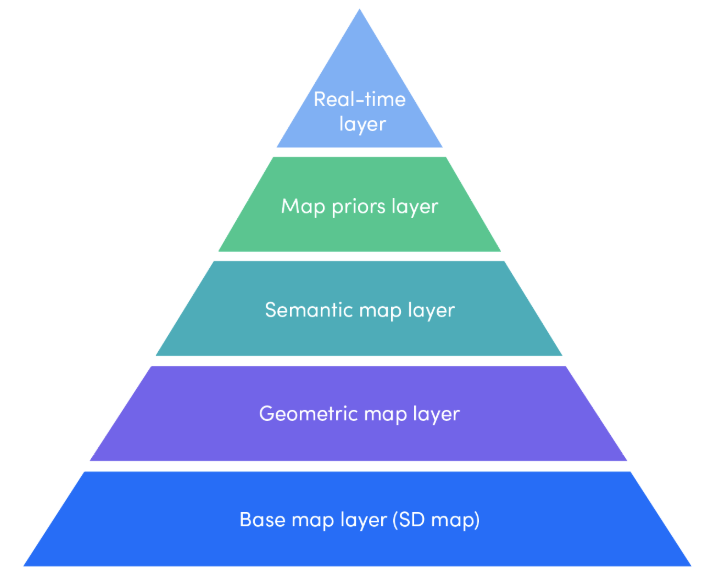}
\centering
\caption[HD-map Layers]{HD-map Layers, Source:{\cite{lyft2}}}
\label{fig:layer_map}
\end{figure}

 However, the implementation of SLAM is not in the scope of this thesis. A pre-built map is generated for the purpose of autonomous navigation. The visualization of HD-maps classified in layers is represented in figure \ref{fig:layer_map}.

For the purpose of this thesis, we will focus mainly on creating a standard definition base map containing 2D geometric information and the semantic map discussed in Chapter \ref{chapter:seg}.

\section{Literature Review}
\subsection{Map Layers}
A map is useless to a robot if it contains no information about its environment. However, by increasing the richness of information in an HD-map leads to an increase in the size of the map. As a result, more processing power would be required to analyze all the information and perform an action based on the information. Hence, what is included in an HD-map and what isn't is decided based on the purpose of the application. For example, a self-driving car would require very high-precision and a detailed map, as it has to take into consideration a lot of factors like the safety of pedestrians, staying in lane, follow traffic rules, etc.
\par The different map layers in an HD-map as mentioned \cite{lyft1}, \cite{lyft2} are:
\begin{itemize}
    \item \hyperref[sm]{Standard Definition Map Layer}
    \item \hyperref[gm]{Geometric Map Layer}
    \item \hyperref[sem]{Semantic Map Layer}
    \item \hyperref[ml]{Map Priors Layer}
\end{itemize}
\subsubsection{Standard Definition Map Layer}
\label{sm}
At the foundation of our map layer is the standard definition map layer. This represents all of the road segments and the interconnections, how many lanes there are, what direction they travel in, and the entire network of roads. It helps the robot to understand the basic attributes of the environment it is navigating in. Although an SD-map contains the basic information about the roads and their basic details, it is not sufficient to smoothly navigate a robot autonomously. The SD-map is used as a base for the other map layers and is used to perform global path-planning for the robot.
\par
\begin{figure}[t]
\includegraphics[width=5.5in]{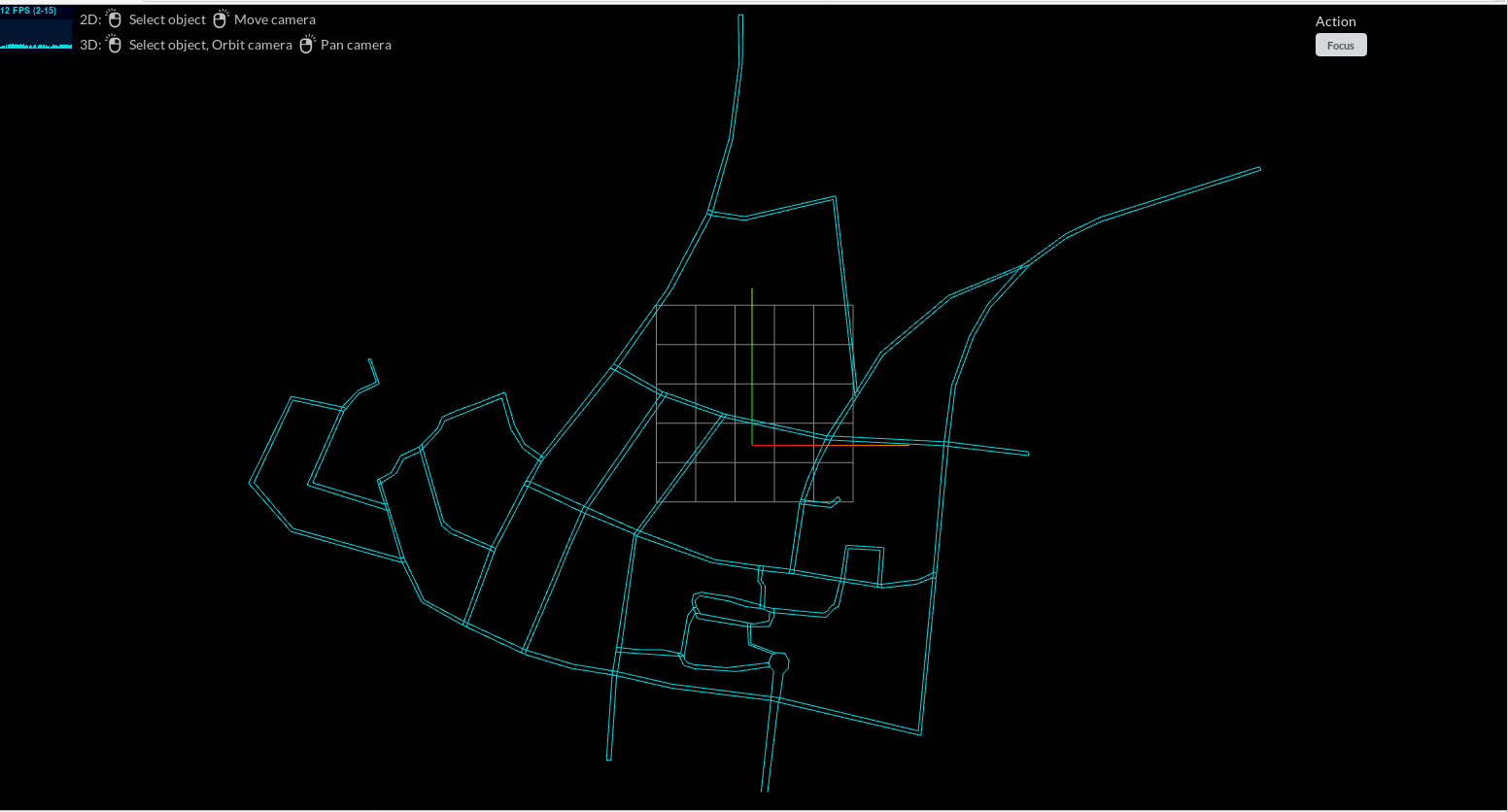}
\centering
\caption{Map of VNIT campus}
\label{fig:map}
\end{figure}

 The SD-map of VNIT campus containing the information about lanes, the direction of the vectors, information about curbs, etc. was mapped using OpenStreetMap(OSM) \cite{OpenStreetMap}. It is an editable map database built and maintained by volunteers and distributed under the Open Data Commons Open Database License. The map created for this project is shown in Figure \ref{fig:map}

\subsubsection{Geometric Map Layer}
\label{gm}
The geometric map layer contains 3D information of the world. This information is organized efficiently to support precise calculations. The 3D map is constructed by fusing the raw information from sensors like Lidar, depth camera, IMU readings, GPS readings, etc. The 3D point cloud achieved is then post-processed to produce the corresponding objects in the geometric map. During real-time processing, the geometric layer is used to access the point cloud information. 
\par However, the sensors used to construct this layer require high precision and good resolution, hence are expensive. Our pipeline uses the information obtained from standard definition maps to localize the robot and for path planning. 
\subsubsection{Semantic Map Layer}
\label{sem}
The semantic map layer builds on the geometric map layer by adding semantic objects. Semantic objects include 2D and 3D objects found in the surroundings such as intersections, lanes, traffic signs, etc. For this project, we have a well-segmented road robust to various lighting conditions as explained in Chapter \ref{chapter:seg}.
\subsubsection{Map Priors Layer}
\label{ml}
The map priors layer contains derived information about dynamic elements. Information here can pertain to both semantic and geometric parts of the map. These priors are used by the prediction and planning systems to determine the behaviour of the objects like traffic lights, the time to spend in a state, etc. and act accordingly. For the case of this project, we are assuming an ideal scenario without taking into consideration the various complexities involved while driving.

\section{Implementation}
\label{map_implementation}
An SD map of VNIT is generated using the software JOSM as shown in Figure \ref{fig:map} containing the features such as the lanes, their direction, curbs, etc. One of the core elements of the OSM data model are the nodes. Nodes are characterized in the data with latitude, longitude and a unique node-id. Some nodes can be assigned special tags in the form of a key-value pair to describe some physical features in the map like building, road, highway, etc. These tagged nodes are used as reference for planning the path to specific locations in the map. The tagged nodes are added in the map manually and the remaining nodes are populated automatically through the JOSM software. The data from the OSM is downloaded in '.xml' format. 
\par To visualize the osm data, we need a visualizer which can interpret the data in the '.xml' format and display the map accordingly with the help of markers. Using the open-source ROS node $osm\_cartography$ from the package $open\_street\_map$, the map can be visualized in the visualizer rviz (rviz is a 3D visualizer for the Robot Operating System (ROS) framework). Simply visualizing the map is not enough. The map is further used by the localization and path planning algorithms and hence its accuracy is of utmost importance.    

\chapter{State Estimation and Localization}
\label{chapter:localization}

\section{Overview}
Localization is the method by which we estimate the state of a robot within the world. In order to be fully functional, a mobile robot must be capable of navigating safely through an unknown environment while simultaneously carrying out the task it has been designed for. For the purpose of autonomous navigation, the robot has to know where it is in the real world with respect to the map, either provided initially or built simultaneously. However, the robot cannot completely rely on the sensors for accurate localization due to the errors inherent in the sensors. For example, a GPS (Global Positioning System) has an error magnitude in metres, an IMU (Inertial Measurement Unit) readings drift over time and its errors accumulate. Hence, such sensors cannot be trusted to give accurate information about the states of the robot. However, by combining the information obtained by various sensors and using probabilistic filters to reduce the errors due to the sensor readings, a better estimate of the states can be obtained. Based on the prior information about the state of the robot, the robot can estimate the current state and localize itself accordingly. The position and orientation can be considered as the state of the robot. An inaccurate localization can lead to system failure, erratic behaviour of the robot and could cause safety hazards to the people around it. Hence, it is of utmost importance to accurately localize the robot in the environment and reduce the errors accumulated due to faulty sensors. 
\section{Literature Review}
The problem of localization can be approached by two methods \cite{amr}:
\begin{enumerate}
    \item Map based Localization - Map is available prior to the process of localization
    \item Simultaneous Localization and Mapping - The pose of a robot and the map of the environment are estimated at the same time.
\end{enumerate}
 For this project, a map based localization approach is adopted.
 \par Generally sensors like a GPS or GNSS are used to estimate the position in the world using the method of trilateration. However, a GPS may have an error from 1 - 10 metres. The errors may be attributed to a number of errors like :
 \begin{enumerate}
     \item Satellite Geometry
    \item Satellite Orbits
    \item Multipath Effect
    \item Atmospheric Effects
    \item Clock Inaccuracies and Rounding Errors
 \end{enumerate}
For the application of an autonomous robot, such errors are not sustainable. Similarly, other sensors attached to the robot like an IMU sensor, Lidar, wheel encoders, vision sensors, etc. give certain information about the states of the robot either directly or indirectly. Every sensor consist of some uncertainty or errors. Accumulation of such errors may cause the robot to behave in an aberrant manner and could cause extreme fatalities while operating alongside humans or deviate from its desired path. Hence, due to the uncertainty in readings, it is impossible to accurately calculate the state of the robot using a deterministic algorithm. However, if the information obtained by all the sensors is fused together and using a stochastic approach, a better estimation of the states could be achieved. This is known as sensor fusion. Sensor fusion can be defined as the combination of sensory data or data derived from disparate sources such that the resulting information has less uncertainty than would be possible when these sources were used individually. 
\par To estimate the states of the robot and reduce the uncertainty in measurements accumulated by the sensors, probabilistic filters are used. Some of the common probabilistic filters used for localization are :
\begin{enumerate}
    \item Bayes Filter \cite{bf} 
    \item Kalman Filter \cite{kf}
    \item Extended Kalman Filter (EKF) \cite{ekf}
    \item Error State EKF (ES-EKF) \cite{esekf}
    \item Unscented Kalman Filter (UKF) \cite{ukf}
\end{enumerate}
\subsection{Probabilistic Estimators}
\subsubsection{Kalman Filter}
\par Kalman Filter is one of the most widely used probabilistic estimator algorithm. It can be used in all the fields where there is an uncertainty in determining the state of a dynamical system. Using a Kalman filter, an educated guess can be made about the state of the system. The Kalman Filter makes use of the 'prediction' and 'correction' cycle iteratively to estimate the state of the system. Kalman
filters are ideal for systems which are continuously changing. They have the advantage that they are light on memory as they don’t need to keep any history other than the previous state, and they are very fast, making them well suited for real time problems and embedded systems.  
\par
\begin{figure}[t]
\includegraphics[width=5.5in]{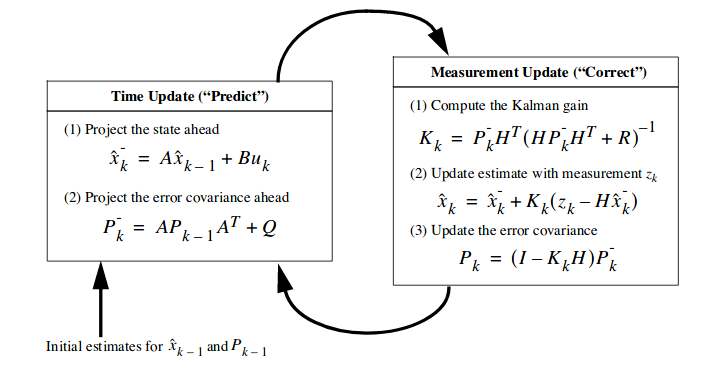}
\centering
\caption{Kalman Filter Equations}Source:\cite{kalman_filter_image}
\label{fig:KF}
\end{figure}

 A Linear Kalman filter is considered to be the best linear unbiased filter. This means, there is no estimator for the state which has a linear state model which is better. It assumes the noise is Gaussian. If the noise is Gaussian, then the Kalman filter minimizes the mean squared error of the estimated state parameters. Two assumptions are taken in the Linear Kalman filter :
\begin{enumerate}
    \item Kalman Filter will always work with Gaussian Distribution.
    \item Kalman Filter will always work with Linear Functions.
\end{enumerate} The algorithm for implementing a Kalman filter is shown in Figure \ref{fig:KF}

\par Unfortunately, systems in real life rarely show linear characteristics. A Kalman filter will give accurate results if the operating range is in the linear zone. Although, quite often, the systems are generally non-linear or have a small range for linear operation. In such cases, a Kalman filter may not be the best choice for an estimator. Hence, even if the Kalman filter is the best linear filter, it is not good enough for a non-linear system. Most real world problems involve non-linear functions and hence we would have to consider a suitable filter accordingly. For this case, a common non-linear filter used is the Extended Kalman Filter (EKF) \cite{ekf}. 

In case of an EKF, the mean of the Gaussian on the non-linear curve is calculated and a number of derivatives are performed to approximate it using the Taylor's Theorem. As the function needs to be linearized, only the first derivative of the Taylor's series is considered. The algorithm of EKF is similar to the Kalman Filter. Hence for this project EKF is to be implemented for the purpose of localization.

\subsubsection{Extended Kalman Filter}
EKF is undoubtedly the most  widely used non-linear estimator techniques that has been applied in the last decade. As mentioned by Lawrence Schwartz and Edwin Stear in \cite{filter}:
 \begin{center}
`It appears that no particular approximate [nonlinear] filter is consistently better than
any other, though ... any nonlinear filter is better than a strictly linear one.'
\end{center} 
EKF is based on linearizing the non-linear functions using the first-order Taylor series expansion. Higher order approaches to non-linear filtering is also possible which provide better results than EKF, but at the expense of greater complexity and computational cost. Hence, EKF is generally used as it is a light-weight non-linear filter as compared to higher order estimators.
\par Consider the following general non-linear system: 
\begin{flalign*}
    \dot{x} &= f(x, u, w, t) \\ 
    y &= h(x, v, t) \\
    w &\sim (0, Q) \\
    v &\sim (0, R)
\end{flalign*}
    The system equation $f(.)$ and measurement equation $h(.)$ are non-linear functions, where $x$ represents the states of the system, $u$ is the input, $w$ is the process or motion noise which is a Gaussian function with zero mean and $Q$ covariance. $v$ is the measurement noise with $R$ covariance. A zero mean white Gaussian noise model is generally taken to mimic the random processes that occur in nature.
    \par
\begin{figure}[t]
\includegraphics[width=3.5in]{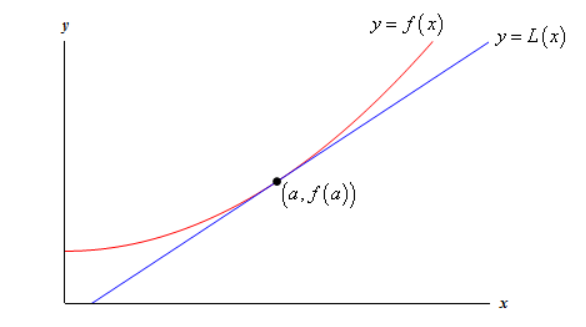}
\centering
\caption{EKF Linearization}Source:\cite{linearize}
\label{fig:ekf}
\end{figure}

 Taylor series is used to linearize the non-linear functions about a nominal control $u_0$, nominal state $x_0$, nominal output $y_0$ and nominal noise values $w_0$ and $v_0$. These nominal values are generally based on a priori guesses of what the system might look like. As mentioned in \cite{non-linear_estimators}, in EKF the Kalman filter estimate is used as the nominal state trajectory. We linearize the nonlinear system around the Kalman filter estimate, and the Kalman filter estimate is based on the linearized system. As shown in figure \ref{fig:ekf}, an operating point `a' is selected and a linear approximation is carried out using the first-order Taylor series.
    
    \begin{flalign*}
        \dot{x} &\approx f(x_0, u_0, w_0, t) + \frac{\partial f}{\partial x}\rvert_{0} (x - x_0) + \frac{\partial f}{\partial u}\rvert_{0} (u - u_0) + \frac{\partial f}{\partial w}\rvert_{0} (w - w_0) \\
         &= f(x_0, u_0, w_0, t) + A\Delta x + B\Delta u + L\Delta w \\
        y &\approx h(x_0, v_0, t) + \frac{\partial f}{\partial x}\rvert_{0} (x - x_0) + \frac{\partial f}{\partial v}\rvert_{0} (v - v_0) \\
        &= h(x_0, v_0, t) + C\Delta x + M\Delta v
    \end{flalign*}

After the linear approximation of the non-linear functions, the Kalman filter equations can be applied to get the estimate as mentioned in \cite{non-linear_estimators}. 

\begin{flalign*}
    \hat{x_0} &= E[x(0)] \\
    P(0) &= E[(x - x_0)(x - x_0)^T] \\
    \dot{\hat{x}} &= f(\hat{x}, u, w_0, t) + K[y - h(\hat{x}, v_0, t)] \\
    K &= PC^T\Tilde{R}^{-1} \\
    \dot{P} &= AP + PA^T + \Tilde{Q} - PC^T\Tilde{R}^{-1}CP \\
    \Tilde{Q} &= LQL^T \\
    \Tilde{R} &= MRM^T
\end{flalign*}

\par  Although the EKF gives better results than a linear Kalman filter, it has several drawbacks as explained in \cite{ekf_limitations}:
\begin{enumerate}
    \item Linearization Error : The difference between the linear approximation and the non-linear function is called linearization error as shown in figure \ref{fig:error}. The linearization errors generally depend on:
    \begin{enumerate}
        \item Non-Linearity of the function
        \item How far away from the operating point the linear approximation is being used
    \end{enumerate}
    
    \begin{figure}[t]
    
    \includegraphics[width=3.5in]{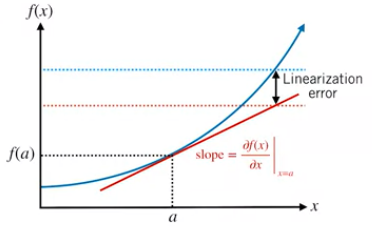}
    \centering
    \caption{Linearization Error}Source:\cite{ekf_limitations}
    \label{fig:error}
    \end{figure}

    \item Computing Jacobians :
    \begin{enumerate}
        \item Analytical differentiation is prone to human error.
        \item Numerical differentiation can be slow and unstable.
        \item Automatic differentiation (e.g., at compile time) can behave unpredictably.
    \end{enumerate}
    \item For highly non-linear functions, the EKF estimate can diverge and become unreliable.
    
\end{enumerate}

There are several other filters like the particle filter, UKF, ES-EKF, etc. which give better performance than the EKF, but at the cost of higher computational requirements. Hence, there is always a trade-off between choosing the filter with better performance and the computational cost associated with it. As EKF is light-weight as compared to other filters and gives sufficiently good enough results, it is widely used as a suitable probabilistic estimator. Hence, for this project, we are going to use an EKF as the estimator used for localization.

\section{Sensor Fusion}
\label{sensor_fusion}

Sensor fusion is the method of combining the measurement readings from different sensors attached on the robot to get a better estimate. As discussed earlier, an autonomous robot cannot depend on the information provided by one or two sensors to localize itself precisely in the environment. Hence by extracting the information from multiple sensors and fusing the data using the probabilistic models explained earlier, we reduce the uncertainty and obtain better results. 
\par The major question that arises is how many sensors are actually essential and how to choose the sensors required? Increasing the number of sensors surely increases the performance, but also increases the computational cost and also the overall cost of the robot. For the purpose of localization, we need a sensor to get the global position estimate, a sensor to determine the orientation of the robot and other sensors to find the odometry of the robot. In this project, we are using a GPS for getting the global position update, an IMU sensor to find the orientation, a 360$^{\circ}$ 2D range scanner (YDLIDAR X4) and a depth camera (Orbbec Astra) for this purpose. The mentioned sensors were chosen for the following reasons \cite{sensor_fusion}:
\begin{enumerate}
    \item The error dynamics are completely different and uncorrelated.
    \item IMU provides smoothing of the GPS readings.
    \item GPS provides absolute positioning information, which reduces the IMU drift.
    \item The depth camera provides accurate local positioning within known maps.
    \item Lidar provides odometry data in places where the camera fails to give output, e.g., in dark conditions, in presence of direct sunlight on the IR receiver, false loop closure, etc.
\end{enumerate}
\subsection{Global Positioning}
\begin{figure}[t]
\includegraphics[width=4.5in]{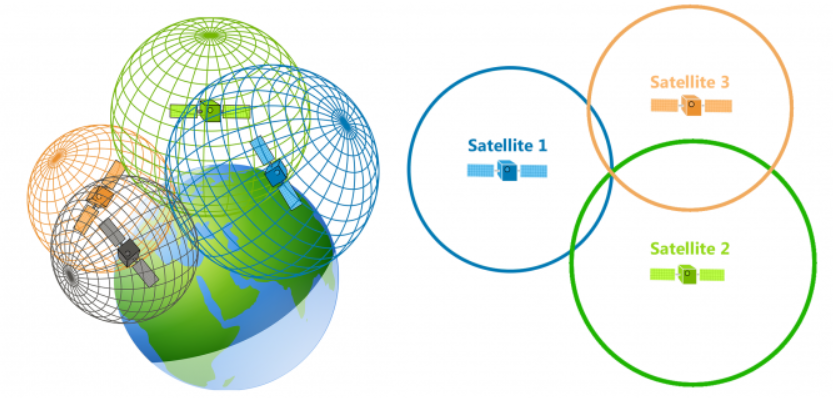}
\centering
\caption{Trilateration}Source:\cite{Trilateration}
\label{fig:gps}
\end{figure}

The Global Positioning System(GPS) is a satellite-based navigation system consisting of a network of 24 orbiting satellites around the earth. The GPS is a US owned utility that provides the users with position, navigation and timing services. In order to obtain the location of a GPS receiver, it uses a process called trilateration. For the process of trilateration to work, at least four satellites must be visible to the location on earth. Each satellite transmits information about its position and current time at regular intervals. These signals, travelling at the speed of light, are intercepted by the GPS receiver, which calculates how far away each satellite is based on how long it took for the messages to arrive. Figure \ref{fig:gps} shows the visualization of the process of trilateration used in a GPS. 

\par For the purpose of an autonomous robot, the GPS plays a crucial role. However, as mentioned before, due to the errors in GPS readings, it can't be completely trusted to give accurate results.  


\subsection{Orientation Estimate}
For the purpose of determining the orientation of the robot, we use an inertial measurement unit(IMU). An IMU is generally present in most of the robots and even smartphones. An IMU has wide scope of applications such as activity tracking, pose estimation, smart-phone applications, gaming, etc. As mentioned in \cite{imu}, a basic IMU mainly consists of:
\begin{enumerate}
    \item Gyroscope which measures the angular rotation rates about three axes. A gyroscope can be considered as a spinning disc that maintains a specific orientation with respect to the inertial space, thus providing an orientation reference as shown in figure \ref{fig:gyro}. Due to recent advancements in the field of Microelectromechanical systems(MEMS), the size and cost of a gyroscope have reduced drastically. However, these systems give noisy readings and the measurements drift over time. The measurement model of gyroscope can be given as :
    \begin{flalign*}
        \omega(t) = \omega_s(t) + b_{gyro}(t) + n_{gyro}(t) 
    \end{flalign*}
    where, $\omega_s(t)$ is the angular velocity of the sensor with respect to the reference frame, $b_{gyro}(t)$ is the gyro bias evolving over time and $n_{gyro}(t)$ is the white Gaussian additive noise term.  

\begin{figure}[h]
\includegraphics[width=1.5in]{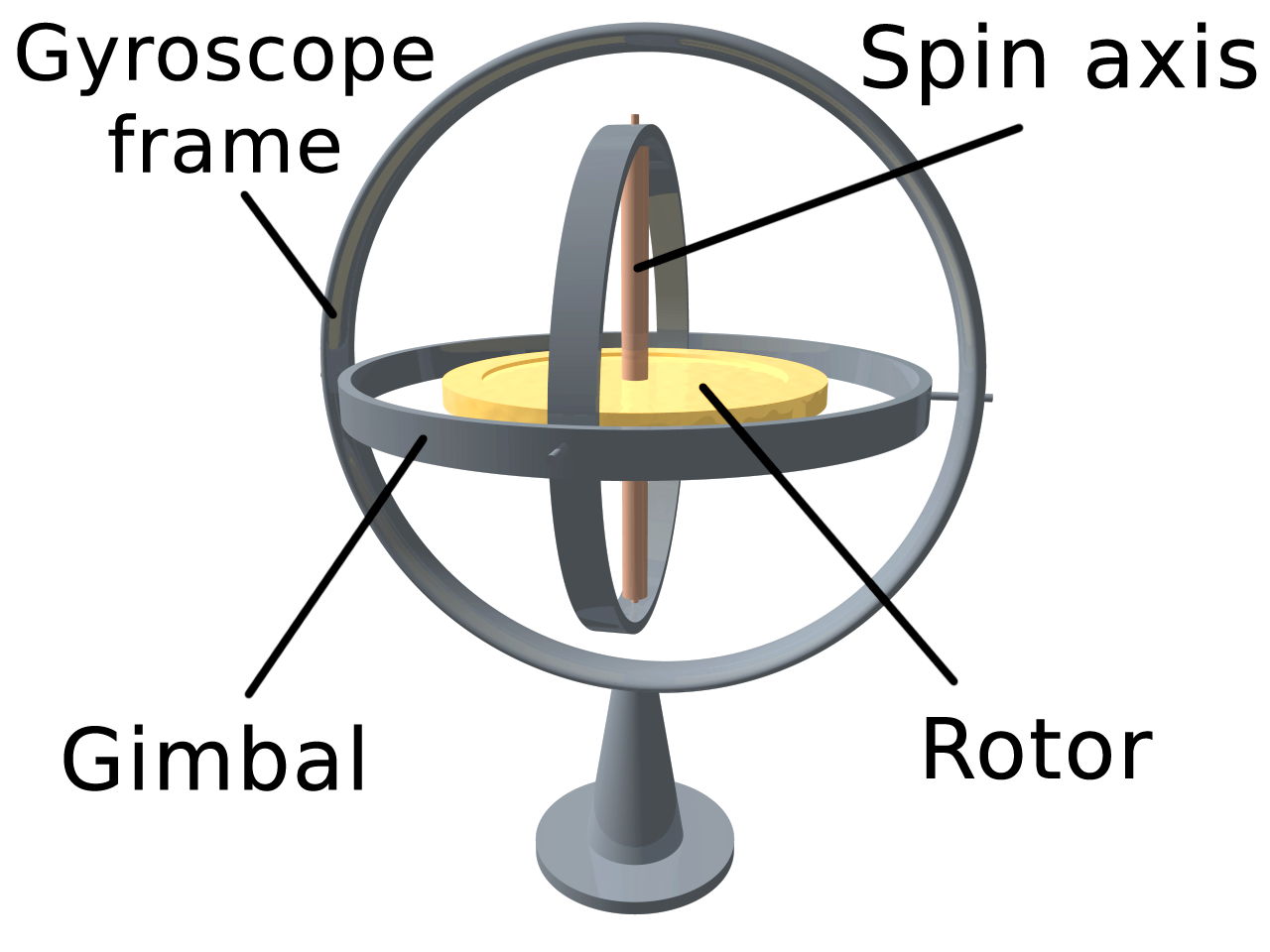}
\centering
\caption{Gyroscope}Source:\cite{Gyro}
\label{fig:gyro}
\end{figure}

    \item Accelerometer which measures the acceleration relative to the gravitational force, also known as specific force. An accelerometer can also be used to measure gravity as a downward force. Integrating acceleration once reveals an estimate for velocity, and integrating again gives you an estimate for position. However, due to the integration, even the errors get accumulated and give erroneous results over time. Hence, this technique of getting the velocity and position estimate using an accelerometer is not recommended.
    \item Some IMUs also contain a magnetometer. It can detect fluctuations in Earth’s magnetic field, by measuring the air’s magnetic flux density at the sensor’s point in space. Through those fluctuations, it finds the vector towards Earth’s magnetic North. Using this data it can be fused with the accelerometer or gyroscope readings to get an absolute heading of the robot. 
\end{enumerate}

Hence, IMU sensor is one of the most important sensor attached on an autonomous robot. An IMU not only gives information about the heading of the robot, but also about the acceleration of the robot which can be fused with other sensors to get an accurate position or velocity estimate. 
\subsection{Visual Odometry}
Odometry is the estimation of the change in the position and orientation over time. The odometry data received from the GPS, IMU and wheel encoders are frequently subjected to mechanical errors, drift, bias, slipping and skidding of the wheels, numerical integration errors, etc. Hence, these sensors are supplemented with visual sensors or laser based sensors. Due to the advancements in the field of computer vision in the past decade, a visual sensor has become an integral part of every robot. Nowadays, a simple camera can be used to perform extremely convoluted tasks like object detection, scene understanding, odometry estimation, etc. and these intricate tasks are performed with very high precision using the computer vision algorithms being developed today. The RTAB-Map(Real-Time Appearance Based Mapping) is a RGB-D, Stereo and Lidar Graph-Based SLAM approach based on an incremental appearance-based loop closure detector. This package is quite robust to provide stable outputs. In this project, the rgbd\_odometry node of the ROS package rtabmap is used which publishes the odom topic used as an input in the estimator. 
\par
\begin{figure}[t]
\includegraphics[width=2.5in]{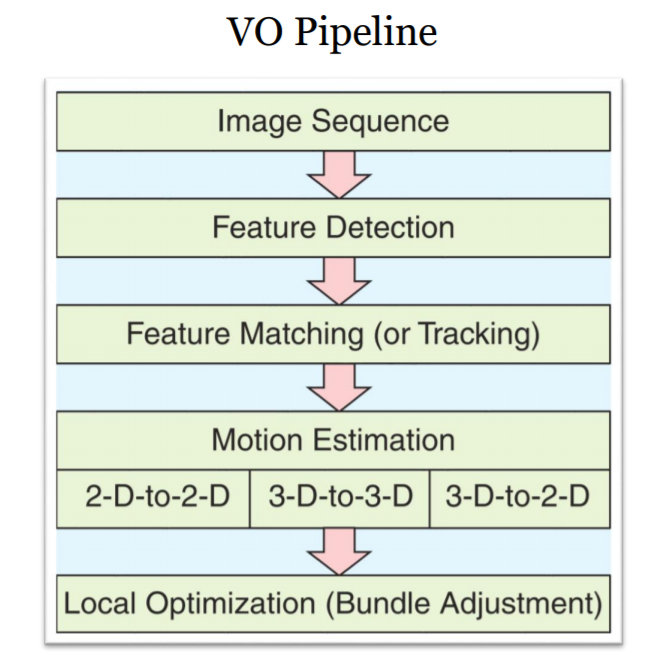}
\centering
\caption{Visual Odometry}Source:\cite{vo_issues}
\label{fig:vo}
\end{figure}

 Visual Odometry makes use of the estimation of the motion of the camera using sequential images i.e ego-motion. The pipeline for visual odomery is depicted in Figure \ref{fig:vo}. A basic algorithm for visual odometry can be explained as mentioned in Algorithm\ref{alg:vo}:

\begin{algorithm}[H]
     \SetAlgoLined
     \caption{Visual Odometry Algorithm}
     
        1. Capture new frame I\textsubscript{k}\\
        2. Extract and match features between I\textsubscript{k-1} and I\textsubscript{k}\\
        3. Compute essential matrix(computed from feature correspondence using epipolar constraint) \\
        4. Decompose essential matrix into R\textsubscript{k} and t\textsubscript{k} and form T\textsubscript{k}\\
        5. Compute relative scale and rescale t\textsubscript{k} accordingly\\ 
        6. Concatenate transformation by computing $ C_k = C_{k-1}T_k $ \\
        7. Repeat from 1
\label{alg:vo}
\end{algorithm}

where, R\textsubscript{k} is the Rotation matrix, t\textsubscript{k} is the Translation matrix and T\textsubscript{k} is the Transformation matrix. 
\par The advancements in the field of computer vision has led to quite accurate means of generating precise odometry information. These methods of obaining the odometry information help in providing a source to mitigate the errors caused due to the gps, imu and other sensors prone to noisy readings. Although, visual odometry is accurate in appropriate lighting conditions, it has several challenges as mentioned in \cite{vo_issues}:
\begin{enumerate}
    \item Robustness to lighting conditions
    \item Lack of features / non-overlapping images
    \item Without loop closure the estimate still drifts
\end{enumerate}

Hence, in order to make our estimator more robust to the challenges mentioned above, we have opted to fuse visual odometry with Lidar odometry.

\subsection{Lidar Odometry}
Lidar(Light Detection and Ranging) sensor is one of the most widely used sensor in autonomous robots and self-driving cars. It has been an enabling technology for autonomous robots to visualize its environment in 360\degree and the Lidar provides very accurate range information. The Lidar uses a simple principle of time-of-flight to generate the point clouds of the environment. A single channel 2D Lidar consists of a laser transmitter. A laser pulse is emitted from the transmitter and on collision with an object, it is reflected back which is then captured by the receiver. The transmitter-receiver system is mounted on a motor rotating in 360\degree. The Lidar consists of an inbuilt timer circuit and an encoder to accurately estimate the time-of-flight of the laser pulses emitted and reflected. Using the time-distance equation of time-of-flight, a point cloud is generated of the environment. In case of multiple channel Lidars, a 3D point cloud can be generated containing precise information about the surrounding. 
\par As the Lidars use laser beams for its operation, they are robust to lighting conditions. This provides a way for autonomous robots to visualize the environment at night or places with low visibility. The high definition maps generated by Lidars are generally used in most of the self-driving cars nowadays. The Lidars not only provide accurate information about the surroundings, the point clouds can be processed to provide valuable information like odometry data, dynamic obstacles, occupancy grid generation, etc.
\par
\begin{figure}[t]
\includegraphics[width=3in]{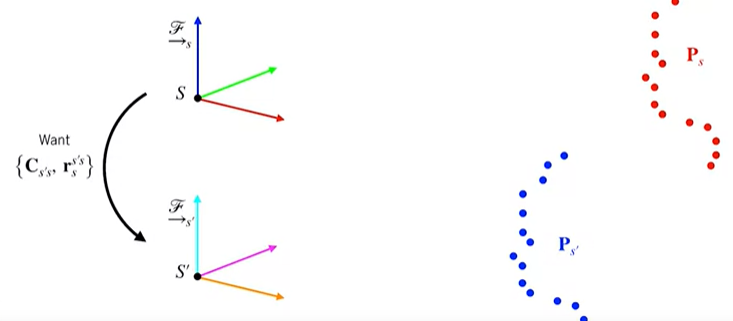}
\centering
\caption{Visual Odometry}Source:\cite{sensor_fusion}
\label{fig:icp}
\end{figure}

 For achieving odometry information from Lidar data, the Iterative Closest Point (ICP) algorithm is most widely used. This problem of finding a spatial transformation to align two point clouds is known as the point cloud registration problem. In the ICP algorithm, one point cloud is fixed as a reference, while the other point cloud(source) is transformed to best match the reference. For example, suppose a reference point cloud is returned by the Lidar at time $t_1$ with respect to a co-ordinate frame $S$. After some time $t_2$ and forward movement of the robot, another point cloud if returned by the Lidar with respect to a co-ordinate frame $S'$. Now, the point set registration problem states that given two point clouds in two different co-ordinate frames, and with the knowledge that they correspond to the same object in the world, how to align them such that the relative motion of the robot can be estimated. As shown in Fig. \ref{fig:icp}, if the correct correspondences are known, the correct relative rotation/translation can be calculated in closed form.

The problem can be formulated as:
\begin{enumerate}
    \item Given: two corresponding point sets:
    \begin{flalign*}
        P_s = \{x_1,..., x_n\}\\
        P_S' = \{p_1,..., p_n\}
    \end{flalign*}
    \item Wanted: translation t and rotation R that
    minimizes the sum of the squared error: 
    \begin{flalign*}
        E(R, t) = \frac{1}{N_p}\sum{i=1}{N_p}\|x_i - Rp_i - t\|^2
    \end{flalign*}
    where $x_i$ and $p_i$ are corresponding points.
\end{enumerate}

The ICP algorithm is explained in Algorithm \ref{alg:icp}:
\begin{algorithm}[H]
     \SetAlgoLined
     \caption{Iterative Closest Point}
     
        1. Get an initial guess for the transformation $\{\check{C}_{S'S}, \check{r}^{S'S}\}$ \\
        2. Associate each point in $P_S'$ with the nearest point in $P_S$\\
        3. Solve for optimal transformation $\{\hat{C}_{S'S}, \hat{r}^{S'S}\}$ \\
        4. Repeat until convergence
\label{alg:icp}
\end{algorithm}     
This is an overview of the point set registration problem which is solved using the ICP algorithm to provide with accurate odometry information. However, the detailed explanation of the algorithm is not in the scope of this thesis. In this project, the icp\_odometry node from the ROS package rtabmap is used to find the odometry from the laser scanner using ICP. 
\par To summarize, by fusing the global position estimate from the GPS, orientation estimates from the IMU sensor, visual odometry and Lidar odometry, we obtain robust localization of the robot. We make use of the EKF probabilistic estimator to fuse the data obtained from the various sensors attached on the robot and to accurately localize the robot in the map and the world. The estimator is robust to comprehend partial failures of the sensors and give continuous readings, so that the robot does not drift with time.
\section{Implementation}
An Extended Kalman Filter(EKF) was used for the task of localization and sensor fusion. The detailed information about the sensors used for the localization process has been described in detail in the literature review. Although, the robot was designed to inculcate the IMU and GPS sensors, it was observed that the cheap sensors gave highly erroneous measurements unable to be solved by sensor fusion. Hence, in order to get a relatively better estimate an android phone was attached on the robot to provide the IMU and GPS readings for testing purpose. The entire pipeline has been implemented in the Robot Operating System(ROS). The following sensors are being used for state-estimation and localization:
\begin{enumerate}
    \item android\_sensors\_driver - For the purpose of getting accurate gps and imu data, an android mobile phone containing the app android\_sensors is mounted on the robot. The app publishes the GPS fixes as sensor\_msgs/NavSatFix and the accelerometer/magnetometer/gyroscope data as sensor\_msgs/Imu.
    \item  Orbbec Astra - Astra is a powerful and reliable 3D camera. It is used for obtaining visual odometry and for semantic segmentation. The technical specifications of Astra camera are given in Table \ref{tab:astra_specs}. Two ROS packages - astra\_camera and astra\_launch, are needed for running the camera on ROS.    
    \item YDLIDAR X4 Lidar - YDLIDAR X4 Lidar is a 360-degree two-dimensional laser range scanner (Lidar). It is used for obtaining the Lidar odometry. A yd\_lidar ROS package is available for operating the Lidar using ROS. The technical specifications of the Lidar are given in Table \ref{tab:lidar_specs}.
\end{enumerate}

\begin{table}[h!]
\caption[Astra Camera Specifications]{Specifications of Astra RGBD Camera}
\centering
\begin{tabular}{llc} 
 \toprule
 \textbf{Sr.No}. & \textbf{Specifications} & \textbf{Technical Details} \\ \midrule
  1. & Range & 0.6m – 8m \\
  2. & FOV & 60°H x 49.5°V x 73°D \\
  3. & RGB Image Res. & 640 x 480 @30fps \\
  4. & Depth Image Res. & 640 x 480 @30fps \\
  5. & Size & 165mm x 30mm x 40mm \\
  \bottomrule
\end{tabular}
\label{tab:astra_specs}
\end{table}

\begin{table}[h!]
\caption[YDLIDAR X4 Specifications]{Specifications of YDLIDAR X4}
\centering
\begin{tabular}{llc} 
 \toprule
 \textbf{Sr.No.} & \textbf{Specifications} & \textbf{Technical Details} \\ \midrule
  1. & Range Frequency & 5000 Hz \\
  2. & Scanning Frequency & 6-12 Hz \\
  3. & Range & 0.12-10 m \\
  4. & Scanning angle & 0-360° \\
  5. & Range resolution & \begin{tabular}{@{}c@{}}< 0.5 mm (Range < 2 m), \\ < 1\% of actual distance (Range > 2 m) \end{tabular} \\
  6. & Angle resolution & 0.48-0.52°\\
  7. & Supply Voltage & 	4.8-5.2 V\\
  \bottomrule
\end{tabular}
\label{tab:lidar_specs}
\end{table}

\subsection{Simulations}
For the purpose of localization, it is important to form a motion model and a measurement model for the estimator to carry out the prediction and correction cycle. For initial simulations, the following assumptions are taken:
\begin{enumerate}
    \item The robot to be equipped with a very simple type of Lidar sensor, which returns range and bearing measurements corresponding to individual landmarks in the environment.
    \item  The global positions of the landmarks are assumed to be known beforehand.  
    \item Known data association, that is, which measurement belong to which landmark.
\end{enumerate}
The data for the simulation are taken from \cite{state} 
The robot motion model receives linear and angular velocity odometry readings as inputs, and outputs the state (i.e., the 2D pose $\begin{bmatrix}x & y & \theta \end{bmatrix}^\intercal$) of the vehicle. The motion model is determined as:
\begin{flalign*}
    x_k = x_{k-1} + T\begin{bmatrix}
cos\theta_{k-1} & 0 \\
sin\theta_{k-1} & 0 \\
0 & 1
\end{bmatrix}(\begin{bmatrix}v_k \\ \omega_k \end{bmatrix} + W_k)
\end{flalign*}

\begin{figure}[t]
\includegraphics[width=3.5in]{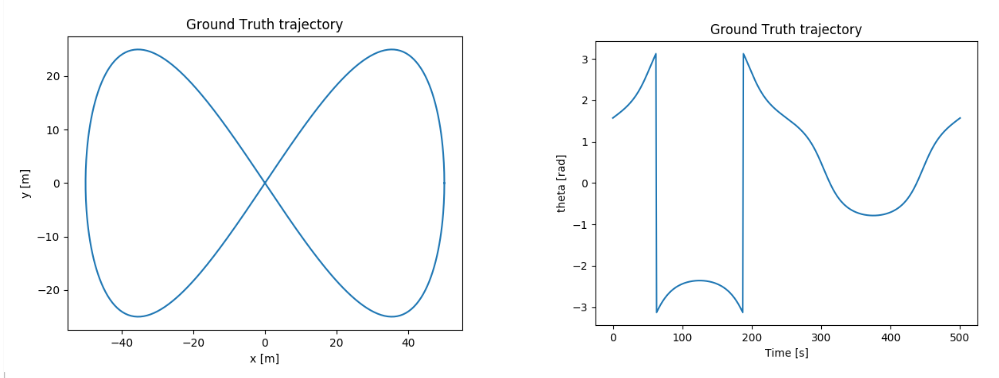}
\centering
\caption{Ground Truth}
\label{fig:gt}
\end{figure}

\begin{figure}[t]
\includegraphics[width=3.5in]{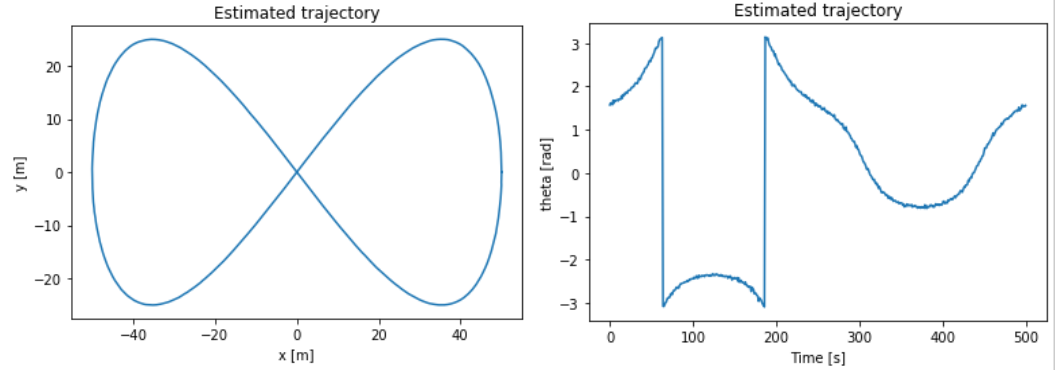}
\centering
\caption{Simulation Result}
\label{fig:sim}
\end{figure}

The measurement model relates the current pose of the robot to the Lidar range and bearing measurements $y_k^l = \begin{bmatrix}r & \phi \end{bmatrix}^\intercal$:
\begin{flalign*}
   y_k^l = \begin{bmatrix}\sqrt{(x_l - x_k - dcos\theta_k)^2 + (y_l - y_k - dsin\theta_k)^2} \\ atan2(y_l - y_k - dsin\theta_k, x_l - x_k - dcos\theta_k) - \theta_k \end{bmatrix} +n_k^l, n_k^l = N(0, R)
\end{flalign*}
$x_l$ and $y_l$ are the ground truth coordinates of the landmark.
$d$ is the known distance between robot center and laser rangefinder (Lidar).
The ground truth data of the trajectory is shown in Fig. \ref{fig:gt} and the simulation results are shown in Fig. \ref{fig:sim}

By tuning the values of the variances, it was observed that:
\begin{enumerate}
    \item One of the most important aspects of designing a filter is determining the input and measurement noise covariance matrices, as well as the initial state and covariance values.
    \item If the sensors are noisy, they will affect the performance of the estimator.
    \item Hence it is necessary to tune the measurement noise variances in order for the filter to perform well.
    \item Improper tuning leads to noisy output as shown in Fig. \ref{fig:noisy}

\begin{figure}[t]
\includegraphics[width=3.5in]{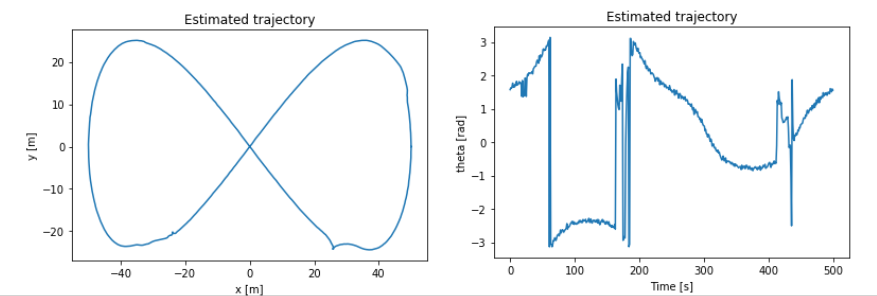}
\centering
\caption{Noisy Result}
\label{fig:noisy}
\end{figure}

\end{enumerate}

\subsection{Tests}
Localization on hardware platform was tested using a ROS package - robot\_pose\_ekf. A couple of indoor and outdoor tests were performed with the robot equipped with all the sensors. Based on the tests, certain conclusions are drawn.      
\subsubsection{Indoor Test}
\begin{figure}[h!]
\includegraphics[width=4.5in]{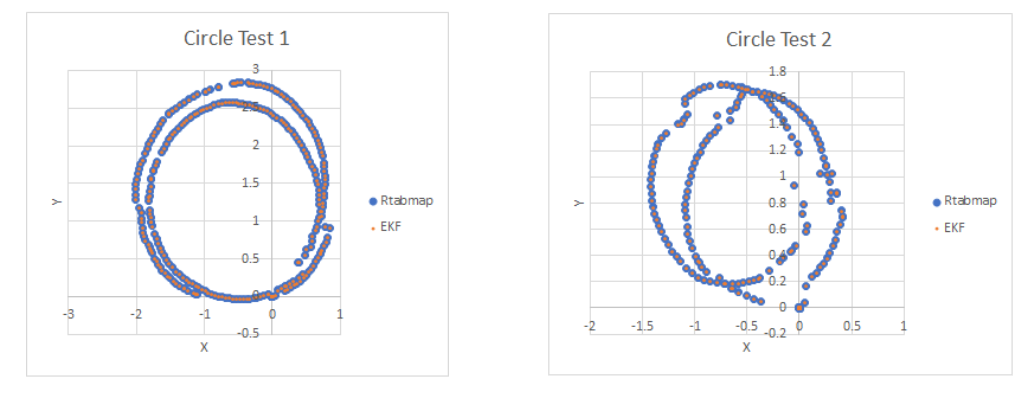}
\centering
\caption{Indoor Circle Test}
\label{fig:indoor}
\end{figure}

For indoor testing of state-estimation, the robot was moved along circular trajectories of different radius. As mentioned in the literature review, in indoor conditions with proper lighting conditions, visual odometry has been proven to be very  accurate. Hence, for the test, the output from visual odometry  is assumed as the ground truth. The results of visual odometry received from rtabmap and the state-estimation received from robot\_pose\_ekf are plotted in Fig. \ref{fig:indoor}

The following things are observed from the indoor tests:
\begin{enumerate}
    \item As shown in Fig. \ref{fig:indoor}, the visual odometry readings are quite accurate as they depict the circle accurately along which the robot is moved in the world.
    \item The readings obtained from the EKF estimator overlap mostly with the ground truth, thus indicating accurate results.
\end{enumerate}

\subsubsection{Outdoor Test}
\begin{figure}[h!]
\includegraphics[width=4.5in]{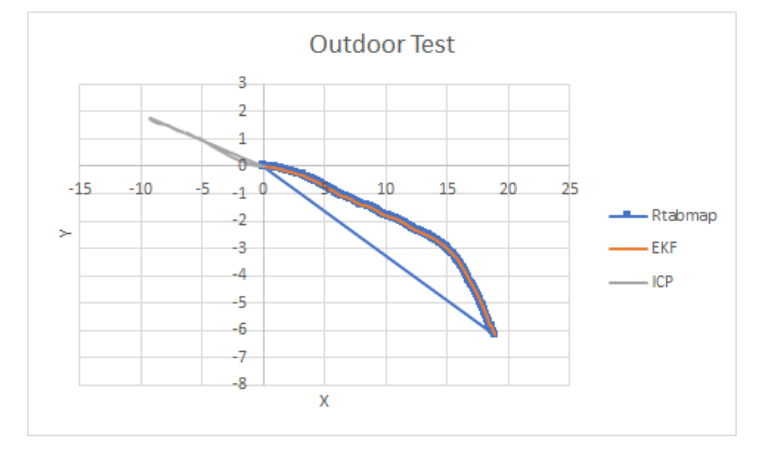}
\centering
\caption{Outdoor Test}
\label{fig:outdoor}
\end{figure}

For outdoor testing, the robot was traversed along a path in the VNIT campus. The sensor data, visual odometry, Lidar odometry received from the ICP algorithm and estimator output were noted for the duration of the test. The results of the test are shown in Fig. \ref{fig:outdoor} 

The following things are observed from the outdoor test:
\begin{enumerate}
    \item As shown in Fig. \ref{fig:outdoor}, in outdoor environment, the data from ICP is not very accurate. This error may be attributed to the sensor noise and false transforms between the sensor frame and odometry frame.
    \item As the test was carried out in daylight, due to proper lighting conditions, the visual odometry data is accurate, but is seen to lose some data. Thus, failing to provide continuous readings.
    \item However, even if the odometry data from visual odometry and Lidar odometry are not continuous and accurate, the EKF estimator is robust enough to provide continuous and precise readings.
\end{enumerate}

\subsection{Conclusion}
 As shown in the Indoor and Outdoor Tests, the following can be concluded:
\begin{enumerate}
    \item In indoor environments, visual odometry gives accurate readings and can be used reliably. However, in the outdoor environments, the visual odometry node tends to fail occasionally, thus providing erroneous readings. 
    \item During tests conducted at night, visual odometry node fails completely due to insufficient inliers and the estimator must depend on Lidar odometry for accurate information.
    \item The robot\_pose\_ekf package fails to provide readings when there is a discrepancy between the timestamps(10 seconds) of the readings from different sensor nodes. This error occurs when two sensor inputs have timestamps that are not synchronized. As sometimes, the visual odometry fails to give feedback due to certain conditions, the readings are not continuous and hence the estimator fails in such cases. 
\end{enumerate}
The importance of accurate state-estimation and localization has been inferred from the various tests carried out on the hardware platform. Localization is the base of any autonomous system. Without precise information about the pose of the robot, it is impossible to function autonomously.
\chapter{Planning Algorithms for Outdoor Environment}
\label{chapter:planning}
\section{Overview}
The planning algorithms decide how the robot will achieve its goal of moving from initial starting point to the destination in an efficient way. The "efficient way" may suggest finding the shortest path, finding the quickest path to reach the goal or finding the path which utilizes the least energy. These constraints are considered in the planning algorithm and an output is a path which the robot can traverse satisfying the required conditions. 
\par Planning at its origin was just a search for a sequence of logical operators or actions that transform an initial world state into a desired goal state. However, the field has flourished to deal with complications such as real world uncertainties, multiple bodies, and dynamics. Presently, planning includes many decision-theoretic ideas such as  imperfect state information, Markov decision processes, and game-theoretic equilibria. Nowadays major research focus in this field is to optimize the process of generating most efficient paths and developing algorithms that unite planning and control.
\par In the context of this project, the planning algorithm needs to decide where the robot should head next. For this purpose, the algorithm takes inputs from perception and localisation units and takes decisions based on our developed procedure. There are three main goals the algorithm needs to achieve: 
\begin{enumerate}
    \item Reaching the desired goal state.
    \item No collisions with other entities(dynamic or static).
    \item Take the best path (shortest or quickest)
\end{enumerate}
\section{Literature Review}
\par
\begin{figure}[h!]
\includegraphics[width=2.5in]{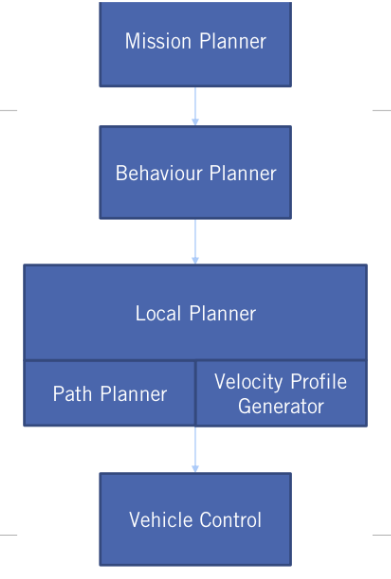}
\centering
\caption{Heirarchical Planning Architecture}Source:\cite{gplan}
\label{fig:heir}
\end{figure}

 The planning architecture is generally  divided into a heirarchical structure as shown in Fig. \ref{fig:heir}

\subsection{Mission Planner}
This is the highest level of optimization problem. In the mission plan, the focus is on map level navigation. The constraints such as finding the shortest path or the quickest path are generally of the major concern here. Other issues such as obstacle avoidance, estimating the time to collision, generating velocity profile, considering the rules of the road etc. are not considered in the mission planner. As mentioned in \cite{gplan}, the mission plan instead focuses on aspects such as speed limits, road length, traffic flow rates, road closures etc. The mission plan is also referred to as Global Plan, as the goal of the mission plan is to find the path the robot will follow in the global map that we generated prior to the operation as mentioned in Chapter \ref{chapter:mapping}.
\par
\begin{figure}[h!]
\includegraphics[width=2.5in]{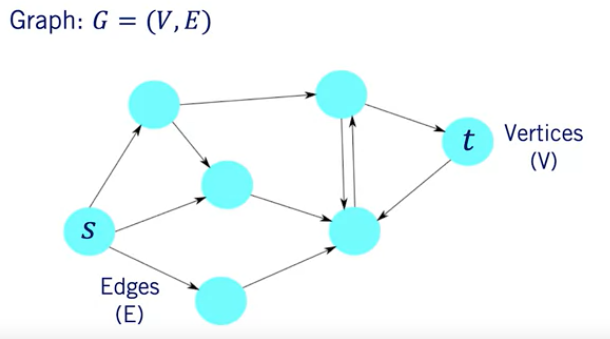}
\centering
\caption{Graph}Source:\cite{graph}
\label{fig:graph}
\end{figure}

 In terms of optimality, the global planner considers the amount of time or distance taken by the chosen path to reach the goal  Graph based algorithms are used to find the path. A graph is a data-structure consisting of vertices and edges $G = f(V, E)$ as shown in Fig. \ref{fig:graph}  

In context to the problem of path planning, the vertices may be considered as nodes in the global map and and the edges are the road networks joining the vertices. In this sense, a contiguous road network can be discretely represented in the form of a graph. Many graph searching algorithms have been developed such as :
\begin{enumerate}
    \item Breadth First Search (BFS)
    \item Depth First Search (DFS)
    \item Djikstra's Algorithm
    \item A-star (A*) Search
    \item RRT, etc.
\end{enumerate}
The BFS and DFS algortihms are basic algorithms which can find the shortest path between the start position and the goal position, but are computationally expensive, not optimal and may get stuck in bug-traps. One of the major drawbacks to the DFS and BFS algorithms are that they do not use weighted edges and thus provide non-optimal solutions. However, the A* search algorithms undertakes a heuristic based approach to find the optimal path. For this thesis, the A* algorithm is implemented for finding the optimal global path.
\subsubsection{A-Star Algorithm}
Before diving in the details of the A* algorithm, certain terminologies are important for understanding the process : 
\begin{enumerate}
    \item agent - An agent is an entity that perceives the environment and acts upon the environment.
    \item state - A state is the configuration of the agent in the environment.
    \item initial state - It is the state where the agent begins its operation.
    \item actions - These are the choices that can be performed in a given state.
    \item transition model - It is a description of what the state results from performing an action in a state.
    \item state space - It is the set of all possible states reachable from the initial state by any sequence of actions.
\end{enumerate}

Unlike the DFS and BFS algorithms, A* algorithm uses weighted edges which may represent the road lengths between two nodes in the mission planner case. To increase the efficiency of the search algorithm, search heuristic is used. In the context of path planning, a search heuristic is an estimate of the remaining cost to reach the destination vertex from any given vertex in the graph. $$h(v) = \|t - v\|$$ However, any heuristic used will not be exact as it would then mean knowing the answer to the problem already. But using the search heuristic helps to find the optimal path faster and prevents getting stuck in bug-traps. 
\par The A* algorithm is explained in Algorithm \ref{alg:A*} \cite{gplan}.

\begin{algorithm}[H]
     \SetAlgoLined
     \caption{A-Star Algorithm}
     
        1. \STATE  open $\leftarrow$ MinHeap()\\
        2. \STATE closed $\leftarrow$ Set()\\
        3. \STATE predecessors $\leftarrow$ Dict() \\
        4. \STATE open.push(s, 0)\\
        5. \While{!open.isEmpty()}{
            u, ucost $\leftarrow$ open.pop() \\
            \If{isGoal(u)}{
              return extractPath(u, predecessors)}
            \For {all v $\in$ u.successors()}{
            \If{v $\in$ closed}{
            continue}
            uvCost $\leftarrow$ edgeCost(G, u, v)\\
            \If {v $\in$ open}{
            \eIf {uCost + uvCost + h(v) < open[v]}{
            open[v] $\leftarrow$ uCost + uvCost + h(v)\\
            costs[v] $\leftarrow$ uCost + uvCost\\
            predecessors[v] $\leftarrow$ u}{
            open.push(v, uCost + uvCost)\\
            costs[v] $\leftarrow$ uCost + uvCost\\
            predecessors[v] $\leftarrow$ u}
            }
            }
            closed.add(u)} 
\label{alg:A*}
\end{algorithm}     

\section{Global Planning Implementation}
For the purpose of this project, A* algorithm was implemented for finding the global path. As mentioned in Chapter \ref{chapter:mapping}, a map of the VNIT campus was constructed using OSM. The osm map existing in the xml file format, consists all the information about the map like the nodes, traversable paths, road lengths, node co-ordinates etc. 
\par
\begin{figure}[h!]
\includegraphics[width=2.5in]{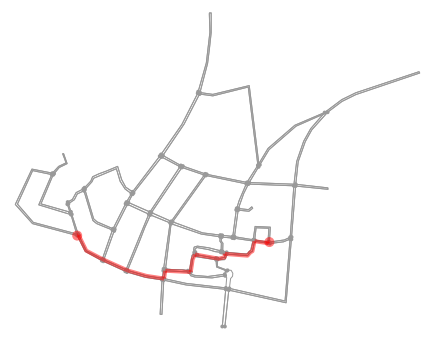}
\centering
\caption{Global Plan}
\label{fig:A*}
\end{figure}

 OSMnx \cite{osmnx} package was used to find the nearest nodes with respect to the nodes in the map and to read the information available in the xml file of the map. Using the A* algorithm explained previously, the shortest path was determined for the robot as shown in Fig. \ref{fig:A*}.

Based on the road network available in the map, an optimal shortest path is obtained through the A* algorithm. This path is just the global map the robot will follow. The global plan is given as an input to the behavioral planner and the local planner. These lower level planners will figure out how to generate obstacle free optimized paths considering the behavior of other dynamic agents in the surrounding, avoiding any collision and also providing a smooth motion to the robot.

\section{Local Planner}
The local planner deals with generating an efficient trajectory and velocity profile for the robot to traverse using the data obtained from other modules of the system. The task of a mission planner or a global plan was to generate a global plan to reach the goal position from the start position as mentioned in the previous section. However, it does not account for the various complexities involved in local motion of the robot. For example, the global plan does not account for generating efficient local trajectories for the robot to follow, it also does not include the information of the immediate surrounding in order to avoid collision with dynamic objects, follow lane, predict the time-to-collision and subsequently avoiding the trajectory leading to collision, etc. The local planner ensures smooth operation of the robot on a well defined, efficient trajectory with proper velocity profile required for the robot to move freely in the environment without any aberration. The local planner is one of the most important part of an autonomous robot as it accounts for the safety of the robot as well as the objects in its surrounding like pedestrians, dynamic vehicles, etc. Slightest of mistake in the local planner can lead to hazardous effects.
\subsection{Basic Terminologies}
Some basic terminologies required before proceeding in this section are mentioned below: 
\begin{enumerate}
    \item Configuration space: A set of all possible configurations of a robot in a given world. It is also called C-space denoted by $C$. The space occupied by a robot is denoted by $A$.
    \item Obstacle space: The space which is already occupied by an obstacle at a given instant of time denoted by $O$. The $C_{obs}$ is defined as a set of all configurations of the robot whose intersection with obstacle space is not null.
    \item Free space: It can be simply defined as $C / C_{obs}$ . That is the space remained by subtracting obstacle space from configuration space.
    \item Path: A continuous function of configurations of robot which will lead the robot to desired state from initial state.
    \item Query: A pair of initial state $(qI)$ and goal state $(qG)$ of robot provided as input to the planning algorithm by human or some other algorithm.
\end{enumerate}

\subsection{Formulation of motion planning problem}
\begin{figure}[h!]
\includegraphics[width=2.5in]{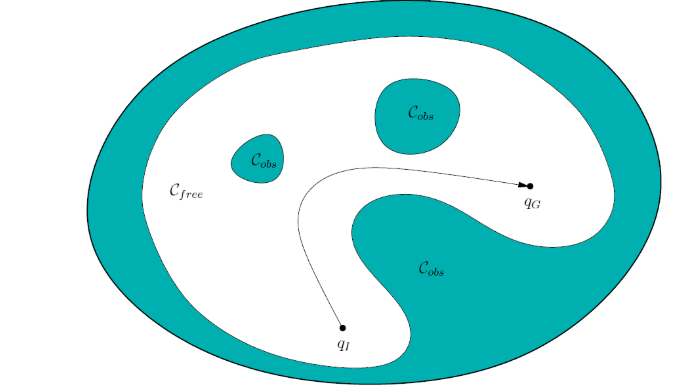}
\centering
\caption{Motion Plan}
\label{fig:plan}
\end{figure}

The basic motion planning problem is conceptually very simple using
C-space ideas. The task is to find a path from $qI$ to $qG$ in $C_{free}$. The entire blob represents $C = C_{free} \cup C_{obs}$. The motion planning problem is shown in Fig. \ref{fig:plan}.

\begin{itemize}
    \item A world $W$ (2D or 3D)
    \item A semi-algebraic obstacle region $O \subset W$ in the world.
    \item A semi-algebraic robot is defined in $W$. It may be a rigid robot $A$.
    \item The configuration space C determined by specifying the set of all possible transformations that can be applied to the robot. From this, $C_{obs}$ and $C_{free}$ are derived.
    \item  A configuration, $qI \in C_{free}$ or the initial configuration.
    \item A configuration $qG \in C_{free}$ or the goal configuration. 
\end{itemize}

A complete motion planning algorithm must compute a (continuous) path, $\tau : [0, 1] \Rightarrow C_{free}$, such that $\tau(0) = qI$ and $\tau(1) = qG$, or correctly deduces that such a path does not exist.
It was shown that this problem is PSPACE-hard, which implies NP-hard.

\subsection{Literature Review}
The currently developed classic methods are variations of four general approaches: Roadmap, Cell Decomposition, Potential fields, and mathematical programming. 

Roadmap approach:
 In this approach, the free C-space, i.e., the set of feasible motions reduced to, or mapped onto a network of 1D lines. This approach is also called the Skeleton, or Highway approach. The search for a solution is limited to the network, and the whole problem becomes a graph-searching problem. 
 \begin{enumerate}
     \item Cell Decomposition: 
In Cell Decomposition (CD) Algorithm, the free C-space is decomposed into a set of simple cells, and then adjacency relationships are computed among the cells. A collision-free path is found by first identifying the two cells containing the initial state and the goal state and then connecting them with a sequence of connected cells.
    \item Potential fields:
This concept was first introduced by Oussama Khatib.In this method, a robot is treated as a point represented in C-space as a charged particle under the influence of an artificial potential field U. The potential function can be defined over free space as the sum of an Attractive potential attracting the robot toward the goal configuration, and a Repulsive potential repelling the robot away from the obstacles to avoid collisions.
    \item The Mathematical programming approach:
This represents the requirement of obstacle avoidance and shortest path with a set of inequalities on the configuration parameters and an objective function. Planning problem is formulated then as a mathematical optimization problem that finds a smooth curve between the start and goal configurations minimizing a certain cost function.
 \end{enumerate}

\subsection{Planning architecture}
The general architecture for planning includes the representation of a pipeline of processing perception inputs to generate optimal plans of motion. It contains several unit operations which are represented in the chart. This gives an overall idea of what’s happening inside the processor.
\subsubsection{Overall Structure}
\begin{figure}[h!]
\includegraphics[width=4.5in]{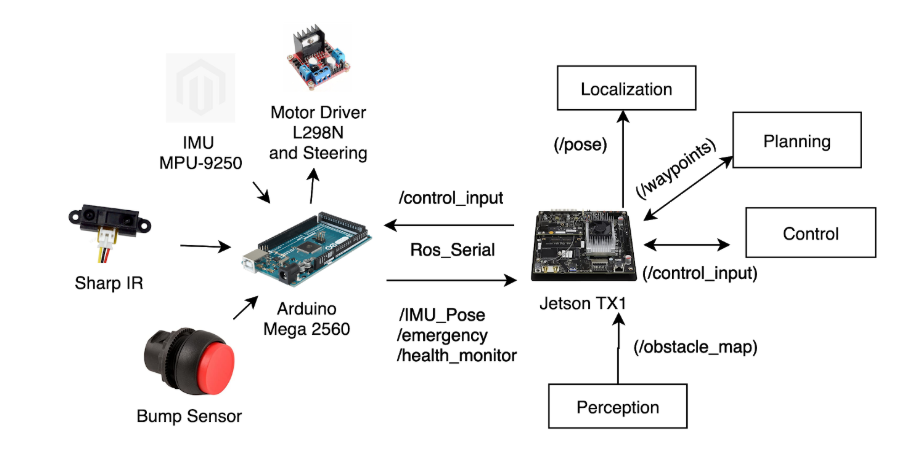}
\centering
\caption{Cyber physical architecture of the robot. Showing all the main processes and the names of ROS topics are mentioned above the arrows to get the intuition of the data transfer. The complete working of each unit is explained throughout the thesis.}
\label{fig:archi_plan}
\end{figure}

First let’s see the entire architecture of the robot. See Fig. \ref{fig:archi_plan}

\begin{figure}[h!]
\includegraphics[width=4.5in]{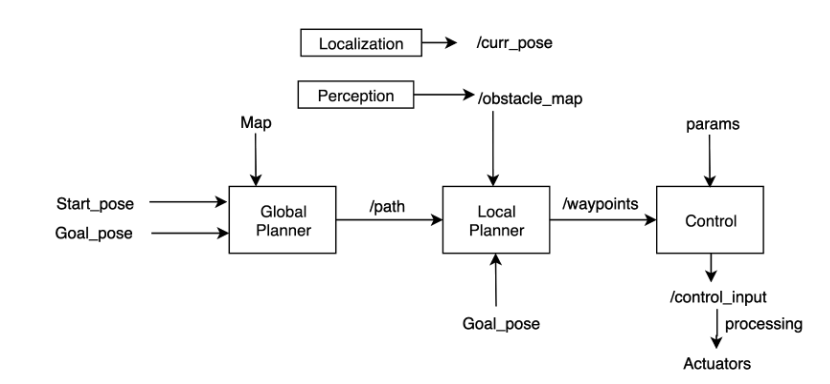}
\centering
\caption{Planning architecture of the robot. The algorithmic details of the above blocks are illustrated in this section. The ROS topic names are shown above arrows to get intuition of the system data transfer.}
\label{fig:planning_archi}
\end{figure}

Now Let’s look at planning architecture. See Fig. \ref{fig:planning_archi}.

\subsubsection{Input processing}
The inputs to the local planner are :
\begin{enumerate}
    \item Global planner:
A-star based searching algorithm which uses a VNIT map to find a path from A to B in terms of nodes which are basically way-points which are a feet away from each other including all the turn nodes. The output of the search is the set of way-points which is published on a ROS topic.
    \item Localisation:
A modified EKF is used to fuse the data from VO, IMU and LIDAR based odom to get a final pose of the robot in the real world. The pose is then transformed into different frames as per requirement. The odometry is published on the ROS topic.
    \item Obstacle data:
A segmentation map generated by a neural net is processed to get the road contour or a boundary of a traversable area on the road. This contour is merged with the cropped lidar contour to get a 360 degree view of the traversable area. The merged contour is published on the ROS topic.

\end{enumerate}

\subsection{Collision checking and avoidance}
\subsubsection{Overview}
\subsubsection{Collision checking}
It is one of the fundamental operations in robotic motion planning. This operation can be divided into static and dynamic collision checking. Static checking refers to checking amounts to testing a single configuration for testing spatial overlaps. Dynamic checking needs to answer if all the configurations on a path in C-space are collision free. There are three major methods for dynamic checking as, Feature tracking , bounding volume and swept volume methods. 
\par 
A common approach is to sample paths at any fixed, pre-specified resolution and statically test each sampled configuration. This approach is not guaranteed to detect collision whenever one occurs, and trying to increase its accuracy by refining the sampling along the entire path results in slow checking. Researchers have found optimal ways to do this by varying sampling resolution as per the requirements in real time. But in our case we have used a fixed resolution to avoid excess computations and we have followed a popular circle based checking method.

\subsubsection{Collision avoidance}
The purpose of obstacle avoidance algorithms is to avoid collisions with the obstacles.These algorithms deal with moving the robot based on the feedback of the sensor information. An obstacle avoidance algorithm is modifying the trajectory of the robot in real time so that the robot can prevent collisions with obstacles detected on the path.

\par We can divide the collision avoidance problem into “global” and “local”. The global techniques involve path planning methods relying on availability of a topological map defining the robots work-space and obstacle space. The entire path from start to goal can be planned, but this method is not suitable for fast collision avoidance due to its complexity. On the other hand the local greedy approaches of using pure obstacle avoidance methods are unable to generate an optimal solution. Another problem is that when using a local approach the robots often get into a local minimum. Because of these shortcomings, a reactive local approach representing obstacle avoidance cannot be considered as a complete solution for  robot navigation. Due to this reason,  we need to combine both obstacle avoidance and path planning techniques to develop a hybrid system overcoming the cons of each of the methods. In our architecture we have also used such a combination to avoid obstacles while planning paths recursively.

\subsubsection{Circle-Point based checking}
Imagine there is a circle $C1$ with center $(x1,y1)$ and radius $D’$. Let’s assume a point $P1$.  Imagine there is a line running between center and $P1$. The distance from the center point to $P1$ when $P1$ is on edge of circle is $D’$ 
\par
\begin{figure}[h!]
\includegraphics[width=4.5in]{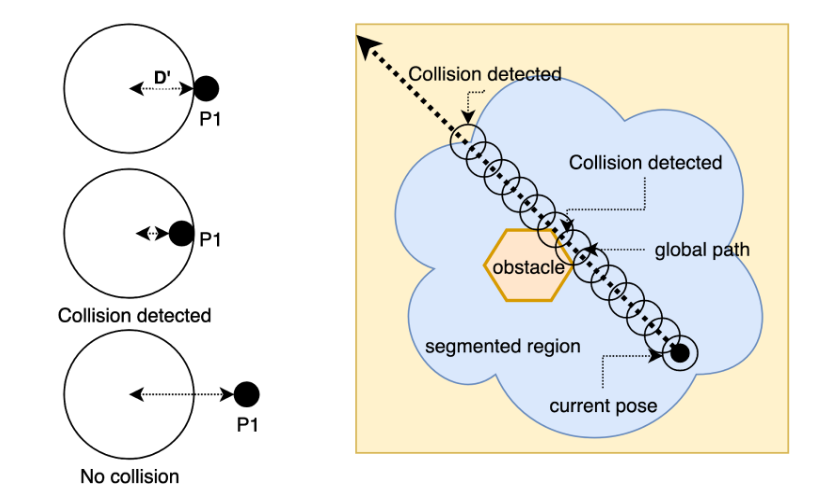}
\centering
\caption{Collision checking method}
\label{fig:col_det}
\end{figure}

So:
Any greater distance than $D’$ and the circle won't collide with the point.
Any less distance than $D’$ and then collision will happen.
The Fig. \ref{fig:col_det} explains it clearly.
We have sampled the obstacle contours into a set of discrete points and we have sampled circles on the global path to check intersection with obstacle contours.

\subsubsection{Safety considerations}
Considering the dimensions of the robot, we have calculated a radius of circle for collision checking which involved an additional cushion along the surroundings of the robot and apart from this the robot is equipped with three sharpIR sensors which are calibrated for a safe distance measurements if the robot goes too close to some obstacle then the emergency stop behaviour is invoked by SharpIR sensors, stopping robot immediately and replanning the route. 
\par The last layer of safety is the bump sensor. If apart from all the algorithmic and hardware provisions, the robot still bumps into something then the bump sensor immediately shuts down the whole system to prevent stalling of actuators and mechanical damage.

\subsection{Local planning algorithm}

\subsubsection{Specific use case}
Considering the environment, motion planning can be either static or dynamic. We say a static environment when the location of all the obstacles is known priori. Environment is dynamic when we have partial information about obstacles prior to robot motion. Initially the path planning in a dynamic environment is done. When the robot follows its path and identifies new obstacles it updates its local map, and changes the trajectory of the path if necessary. In our case the environment is dynamic. Also, our robot is supposed to be travelling along roadsides. On roadsides there might also be some parked vehicles or other static obstacles which needs to be considered while designing planning algorithms. Other than static obstacles there can be humans and other moving vehicles. So we need to be quick and accurate in planning our path. It means that the algo should be computationally efficient, robust and should have predefined emergency behaviours. This defines our use case, now in the next subsection we have mentioned the actual working of the algorithm.

\subsubsection{Designed algorithm}
\begin{figure}[h!]
\includegraphics[width=6.5in]{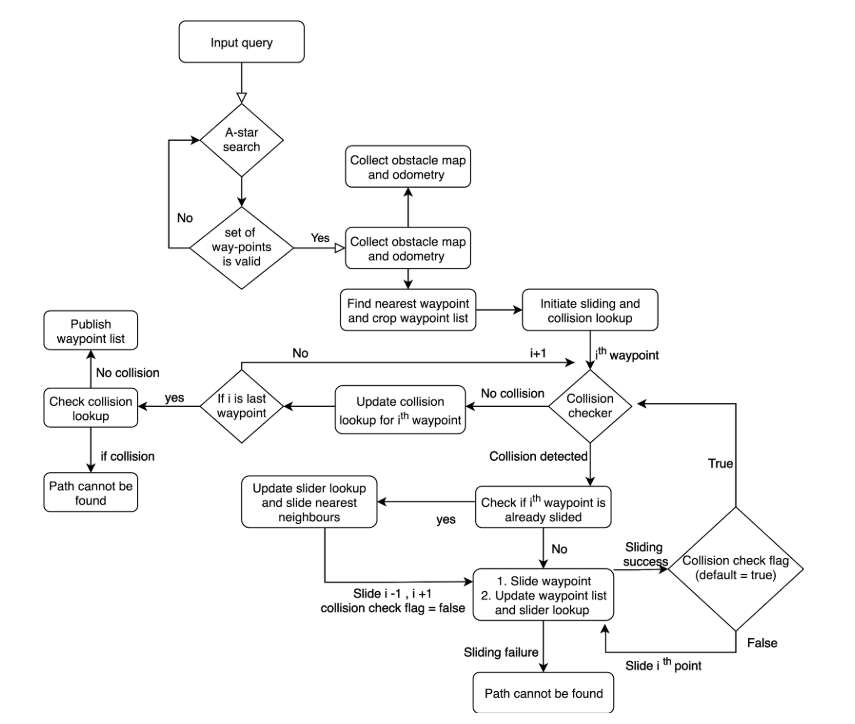}
\centering
\caption{Planning algorithm block diagram}
\label{fig:full_plan}
\end{figure}

\begin{figure}[h!]
\includegraphics[width=4.5in]{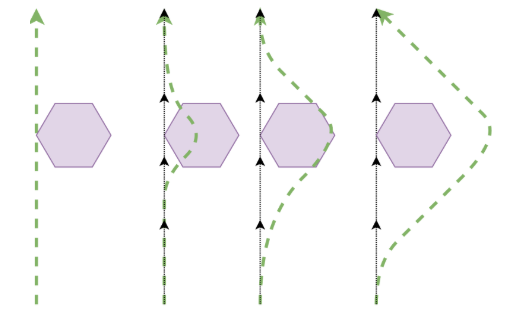}
\centering
\caption{Planning algorithm working explanation. The hexagon indicates an obstacle. The figure is showing a step-wise algorithm when going from the left side of an obstacle is not allowed. The dotted line indicates the path to be followed and the line with the arrow indicates the global path.
}
\label{fig:plan_eg}
\end{figure}

Planner receives the input of obstacle way-points from segmentation contour and Lidar contour.
Way-points are received as input from global planner which is A* based planner searching nodes in a map of VNIT. If collision is detected then way-points are slided  along a line perpendicular to the slope of the line joining the consecutive way-points or path and the direction of sliding is away from obstacles. The distance for sliding is fixed and decided by using dimensions of bot and max turning radius. Sliding is operated like a chain. Every second slide of a way-point is followed by sliding of its immediate next and previous way-points.The collision checker function and avoiding function operate in recursion. The planner returns a new set of way-points to avoid the obstacle. This algorithm will surely return a path if it exists otherwise it reports that a path is not possible. Step-wise algorithm is mentioned in the Fig. \ref{fig:full_plan}. The Fig. \ref{fig:plan_eg} shows an example of how the planner works if sliding to the left slide is not an option.

\section{Conclusion and Future Work}
This thesis explains the hierarchical planning structure and the implementation of A* algorithm to find the shortest path from starting position to the goal in the map along with the local planner responsible for creating smooth and efficient trajectory for the robot with collision avoidance. Further research could be carried out for the optimization of problems related to occupancy grid generation, finding the time to collision, generating energy efficient paths etc.  
\chapter{Semantic Segmentation for Road and Obstacle Detection}
\label{chapter:seg}
\section{Overview}
Image Segmentation is the partitioning of an input image into multiple segments (sets of pixels). The goal is to represent the image as something simpler to analyze and more meaningful. As per \cite{seg_review1}, some of the subtasks involved in image segmentation are :
\begin{itemize}
    \item Semantic Segmentation \cite{semseg_review}: Each pixel is classified into one of the predefined set of classes such that pixels belonging to the same class belongs to a unique semantic entity in the image. Note that the semantics (logic) in question depends not only on the data but also the problem being addressed.
    \item Saliency Detection \cite{salient_objectdet_review}: Focus on the most important object in a scene.
    \item Instance Segmentation \cite{instance_seg}: Segments multiple instances of the same object in a scene.
    \item Segmentation in the temporal space \cite{temporal_seg}: Object tracking requires segmentation in the spatial domain as well as over time (temporal domain).
    \item Oversegmentation \cite{overseg} \cite{color_texture_seg}: Images are divided into extremely small regions to ensure boundary adherence, at the cost of creating a lot of spurious edges. Region merging techniques are used to perform image segmentation.
    \item Color or texture segmentation: Also found to be useful for certain applications.
\end{itemize}
This work focuses on semantic segmentation because that is sufficient for our purpose of identifying the road and obstacles in traffic/road scenes.

\textbf{Note}: Some prior knowledge of Convolutional Neural Networks \cite{deep_cnn_alexnet} is required for proper understanding of this chapter.

\section{Literature Review}
\begin{figure}[t]
\includegraphics[width=5.5in]{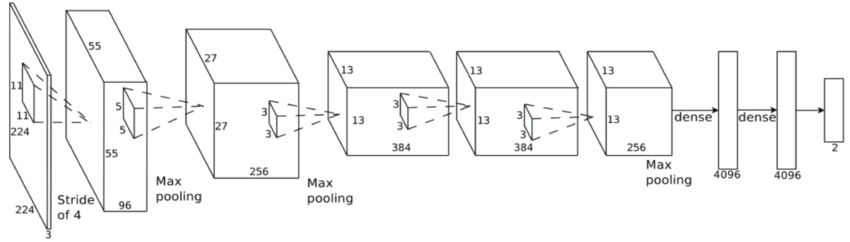}
\centering
\caption[AlexNet Architecture]{AlexNet: Deep Convolutional Neural Network architecture from \cite{deep_cnn_alexnet}}
\label{fig:alexnet}
\end{figure}

Most modern techniques make use of Supervised Deep Learning, majorly involving Convolutional Neural Networks \cite{deep_cnn_alexnet}, due to their success in Image Classification tasks. These convolutional networks consist of sequential application of convolutional filters, pooling layers and non-linear activation functions. An example is shown in Figure \ref{fig:alexnet}. This particular architecture is used for image classification. It is a mapping (function) between the image and the output class.

\begin{figure}[t]
\includegraphics[width=5.5in]{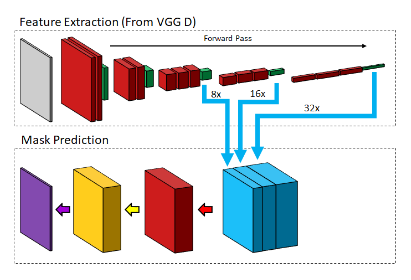}
\centering
\caption[Fully Convolutional Network Architecture]{Fully Convolutional Network architecture from \cite{seg_review2}}
\label{fig:fcn}
\end{figure}

However, semantic segmentation is different, and thus modifications are required to the architecture. Particularly, Fully Convolutional Networks (FCN) \cite{fcn_orig} are used for segmentation. These do not involve fully connected (dense) layers. Further, output size reduces in conventional CNNs, so it cannot be used for pixel-level classification. In FCN, an interpolation layer is used which upsamples the intermediate outputs (called feature maps) to the size of the input image. It is interesting to note that this interpolation is done using bilinear interpolation and is not learnable. An example is shown in Figure \ref{fig:fcn}. 

Fully connected networks (conventional deep CNNs like Figure \ref{fig:alexnet}) used flattening of 2D feature maps to perform classification. However, flattening results in loss of spatial relation between pixels in the feature map. FCNs overcome this by avoiding the use of flattening and using upsampling followed by pixel-level classification. However, the issue with FCNs is the loss of sharpness due to intermediate subsampling. This issue has been addressed using skip connections and other methods.

Another approach is the use of Convolutional Autoencoders \cite{conv_ae}. These are traditionally used for Representation Learning. An autoencoder has 2 parts, encoder and decoder. Encoder encodes the raw input to a lower dimensional representation, while the decoder attempts to reconstruct the input from the encoded representation. The decoder’s generative nature can be modified to achieve segmentation tasks.

The major benefit of these approaches is generation of sharper boundaries without much complication. Unlike classification approaches, the decoder’s generative nature can learn to generate delicate boundaries using the extracted features. Another benefit is that it does not restrict input size. The commonly used technique for decoding is transposed convolutions (learnable) or unpooling layers. However, a possible issue with such approaches is over-abstraction of images during the encoding process, i.e. the network starts memorizing the training images instead of learning filters that are useful for compression and reconstruction.

\begin{figure}[t]
\includegraphics[width=5.5in]{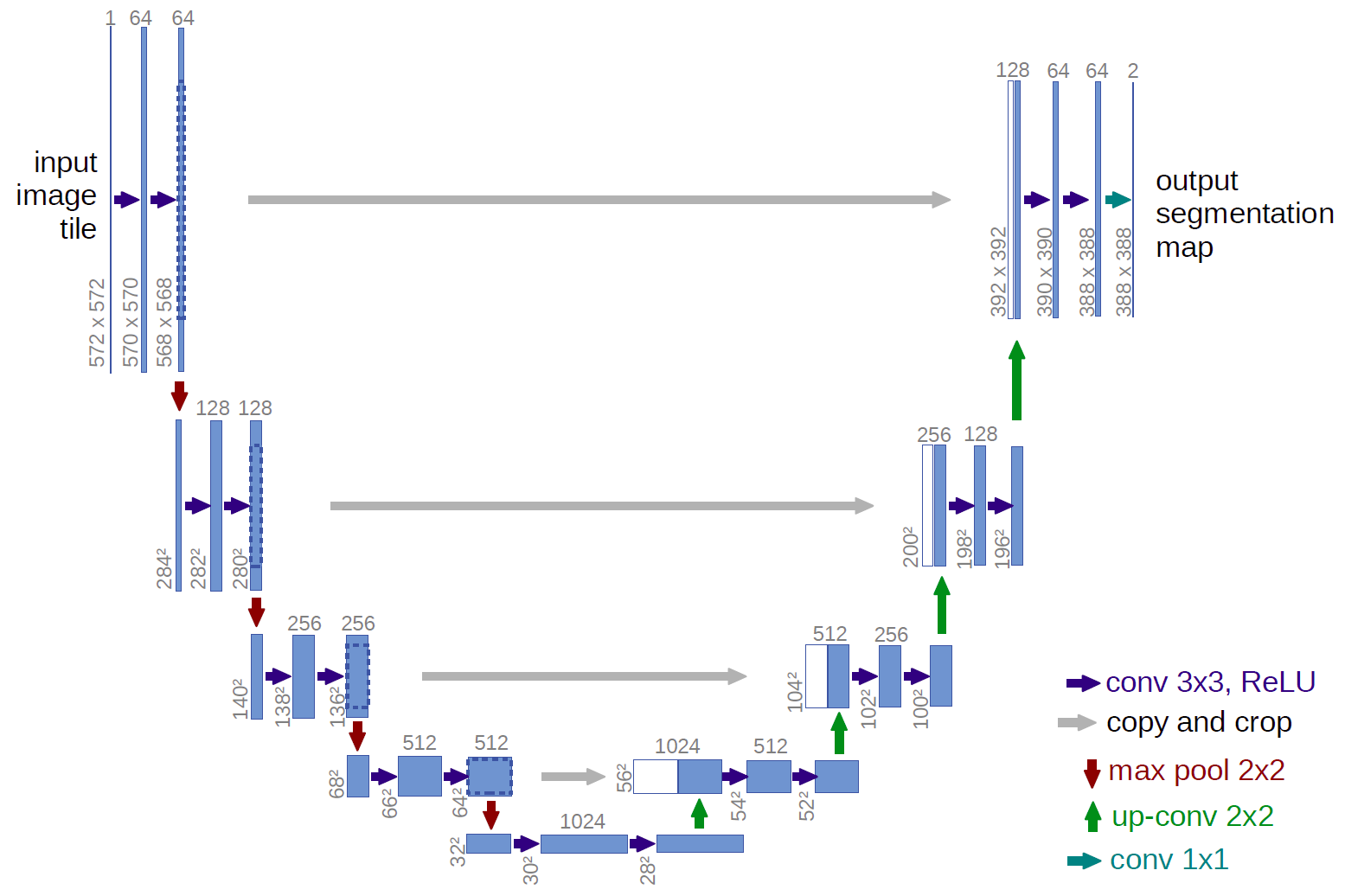}
\centering
\caption[U-Net Architecture]{U-Net architecture utilizing skip connections from \cite{unet}}
\label{fig:unet}
\end{figure}

Another technique is the use of skip connections, first introduced in \cite{resnet}. Linear skip connections are often used to improve gradient flow for large number of layers. Skip connections are also useful to combine different levels of abstraction from different layers to produce sharp segmentation output. An example is shown in Figure \ref{fig:unet}.

\section{Preliminary Experiments}
\begin{figure}[t]
\includegraphics[width=5.5in]{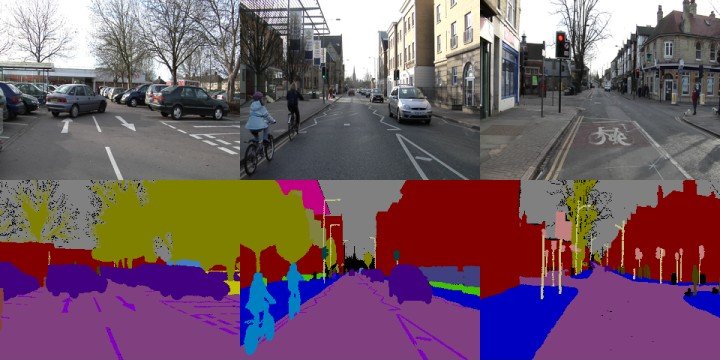}
\centering
\caption[CamVid sample images and labels]{Sample images and pixel-level labels from CamVid \cite{camvid}}
\label{fig:camvid_sample}
\end{figure}

\begin{figure}[t]
\includegraphics[width=5.5in]{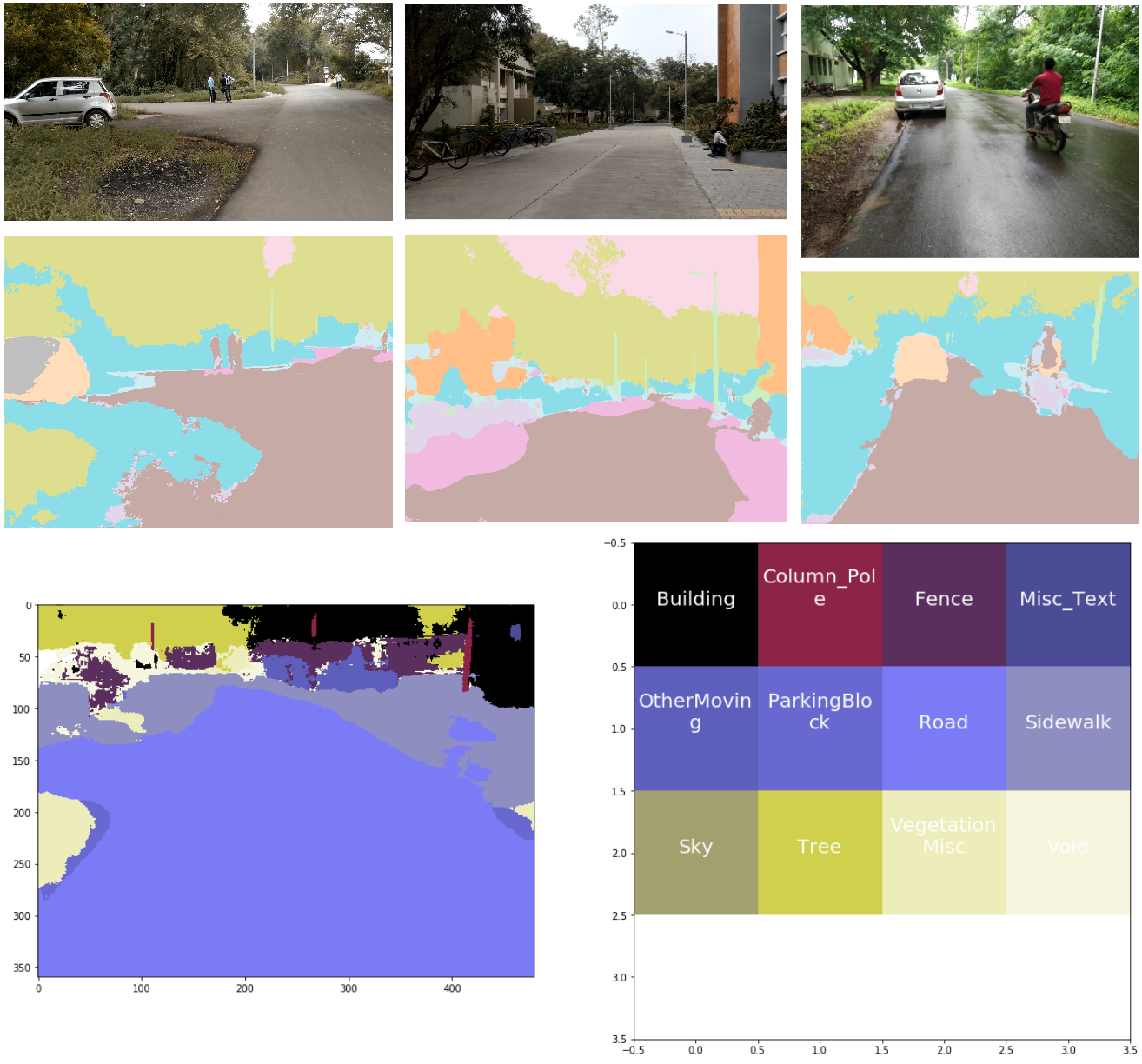}
\centering
\caption[U-Net Results]{Results of U-Net trained with CamVid dataset on VNIT campus images with color map indicating relation between class and color.}
\label{fig:prelim_results}
\end{figure}

For our early tests, we used U-Net architecture \cite{unet} with the Cambridge-driving Labeled Video Database (CamVid) \cite{camvid}. The training is done using Adam Optimizer \cite{adam} with learning rate 1e-4 minimizing pixel-wise Cross Entropy Loss. The implementation was done on the PyTorch \cite{pytorch} based fastai \cite{fastai} framework. The architecture is shown in Figure \ref{fig:unet}. Some example images and labels are shown in Figure \ref{fig:camvid_sample}. The preliminary results are shown in Figure \ref{fig:prelim_results}.

As seen from the results, the output segmentation is not sharp enough and sometimes cannot segment obstacles properly. The major issues which led to this are as follows:
\begin{itemize}
    \item Since the CamVid dataset has only around 700 images, the network is not able to generalize. This also causes the network to inaccurately segment certain obstacles like a person on a motorcycle as in Figure \ref{fig:prelim_results}.
    \item Further, since those images are from urban areas of foreign cities, the network is unable to sharply segment images from our institute's campus, which features less structured roads and other features.
    \item There are too many classes which makes the task more difficult for the network. This is because the underlying optimization problem is more difficult to solve.
\end{itemize}

Another issue, unrelated to the performance, is the high computational complexity of the model. The computational complexity hinders the real-time use of the model, since computational power on the robot will be limited. Thus, for our final implementation, we use ENet \cite{enet}, which is computationally more efficient while preserving performance.

\begin{figure}[t]
\includegraphics[width=5.5in]{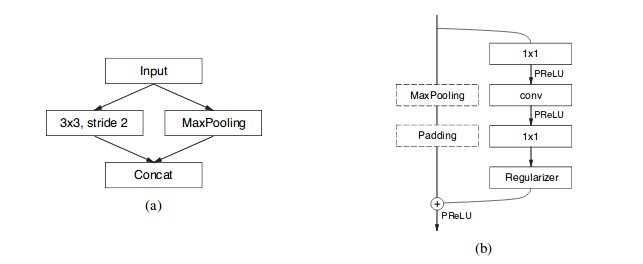}
\centering
\caption[ENet initial block and bottleneck module]{(a) ENet initial block. MaxPooling is performed using non-overlapping 2 $\times$ 2 windows, and the convolution has 13 filters, which sums up to 16 feature maps after concatenation. (b) ENet bottleneck module. conv is either a regular, dilated or deconvolution with 3 $\times$ 3 filters, or a 5 $\times$ 5 convolution decomposed into 2 asymmetric ones. Source: \cite{enet}}
\label{fig:enet_blocks}
\end{figure}

\begin{table}[t]
\caption[ENet Architecture]{ENet Architecture \cite{enet}. Output sizes are given for an example input of 512 $\times$ 512}
\centering
\begin{tabular}{llc}
\toprule
\textbf{Name} & \textbf{Type} & \textbf{Output Size} \\ \midrule
initial &  & $16\times256\times256$ \\ \midrule
bottleneck1.0 & downsampling & $64\times128\times128$ \\ 
4$\times$bottleneck1.x &  & $64\times128\times128$ \\ \midrule
bottleneck2.0 & downsampling & $128\times64\times64$ \\ 
bottleneck2.1 &  & $128\times64\times64$ \\ 
bottleneck2.2 & dilated 2 & $128\times64\times64$ \\ 
bottleneck2.3 & asymmetric 5 & $128\times64\times64$ \\ 
bottleneck2.4 & dilated 4 & $128\times64\times64$ \\ 
bottleneck2.5 &  & $128\times64\times64$ \\ 
bottleneck2.6 & dilated 8 & $128\times64\times64$ \\ 
bottleneck2.7 & asymmetric 5 & $128\times64\times64$ \\ 
bottleneck2.8 & dilated 16 & $128\times64\times64$ \\ \midrule
\multicolumn{3}{c}{Repeat section 2, without bottleneck2.0} \\ \midrule
bottleneck4.0 & upsampling & $64\times128\times128$ \\ 
bottleneck4.1 &  & $64\times128\times128$ \\ 
bottleneck4.2 &  & $64\times128\times128$ \\ \midrule
bottleneck5.0 & upsampling & $16\times256\times256$ \\ 
bottleneck5.1 &  & $16\times256\times256$ \\ \midrule
fullconv &  & $C\times512\times512$ \\ \bottomrule
\end{tabular}
\label{tab:enet_arch}
\end{table}

\section{Final Implementation}
\begin{figure}[t]
\includegraphics[width=5.5in]{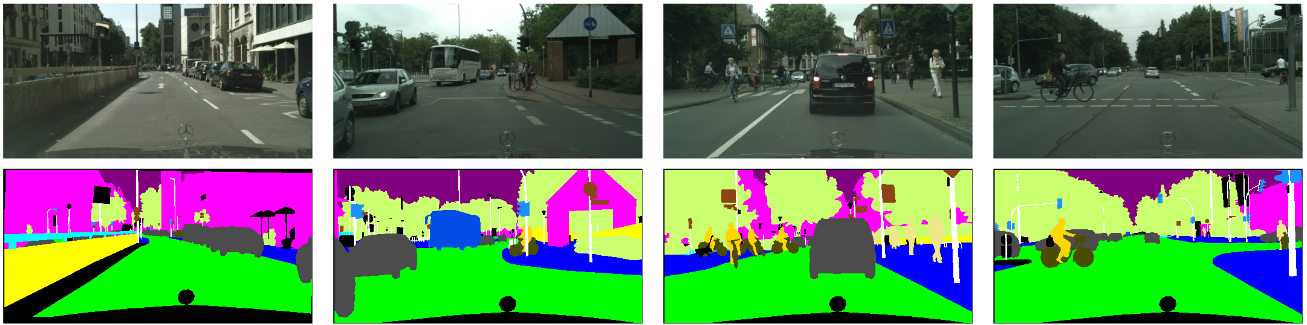}
\centering
\caption[Cityscapes sample images and labels]{Sample images and pixel-level labels from Cityscapes \cite{cityscapes}}
\label{fig:cityscapes_examples}
\end{figure}

E-Net architecture (in Figure \ref{fig:enet_blocks} and Table \ref{tab:enet_arch}) is implemented in PyTorch\cite{pytorch}. The code is available at {\small \url{https://github.com/IvLabs/autonomous-delivery-robot/tree/master/sem_seg_pytorch}}. The dataset used was Cityscapes \cite{cityscapes}, which provides around 25k images having coarse annotations for road scenes. Some examples are shown in Figure \ref{fig:cityscapes_examples}. In this final implementation, only the road is segmented from the image, while assuming anything other than the road to be an obstacle. Further, Dice Loss \cite{dice_loss} is used in combination with the Cross Entropy Loss to improve the segmentation output. Dice Loss is given by: 
\begin{align}
DL(p, \hat{p}) = 1 - \frac{2p\hat{p} + 1}{p + \hat{p} + 1}
\end{align}
where $p \in {0, 1}$ and $0 \leq \hat{p} \leq 1$, $p$ is the label and $\hat{p}$ is the prediction from the network.

Some of the features of this loss function are: 
\begin{itemize}
    \item The fraction $\frac{2p\hat{p} + 1}{p + \hat{p} + 1}$ signifies the amount of overlap between the label and the predicted segmentation. It is called the Dice Coefficient.
    \item Maximizing the Dice Coefficient will give maximum overlap. However, we are already minimizing the Cross Entropy Loss. Thus, we minimize $1 - \text{Dice Coefficient}$, which is the same as maximizing the Dice Coefficient.
    \item Further, this loss function performs particularly well in case of unbalanced segment sizes i.e. very small obstacle segments may affect the optimization and it may get stuck in a local minima. Dice Loss helps to avoid this problem.
    \item However, the problem with Dice Loss is that the gradients are complicated compared to Cross Entropy gradients and have the tendency to explode (too small values of $p$ and $\hat{p}$ may cause very large gradients), which makes the training unstable.
\end{itemize}

Thus, taking note of the advantages and disadvantages of Dice Loss, it is beneficial to use a combination of Dice Loss and Cross Entropy Loss. This combination is minimized simultaneously in our implementation using Adam optimizer with a learning rate of 1e-4. 

The design choices of the ENet architecture are detailed in \cite{enet}. They are briefly summarized below:
\begin{itemize}
    \item The use of \textbf{P}arametrized \textbf{Re}ctified \textbf{L}inear \textbf{U}nits (PReLU) \cite{prelu} gives an additional learnable parameter to the non-linear nature of the network. This improves the performance over using standard ReLU non-linearity.
    \item In other architectures, decoder is generally an exact mirror of the encoder (w.r.t. architecture). Here, decoder is much smaller than the encoder. The idea is that encoder will process and filter the input, while the decoder simply upsamples and fine-tunes the encoder output.
    \item The use of factorizing filters or asymmetric convolutions \cite{asymmetric_conv} i.e. decomposition of $n \times n$ convolution into an $n \times 1$ convolution followed by a $1 \times n$ convolution reduces the computational and memory costs. For example, an asymmetric convolution with $n = 5$ is similar to a $3 \times 3$ convolution in terms of computational cost and memory requirements.
    \item The use of dilated convolutions \cite{dilated_conv} gives the network a wide receptive field. Thus, dilated convolutions are used instead of normal convolutions.
\end{itemize}

\begin{figure}[t]
\includegraphics[width=5.5in]{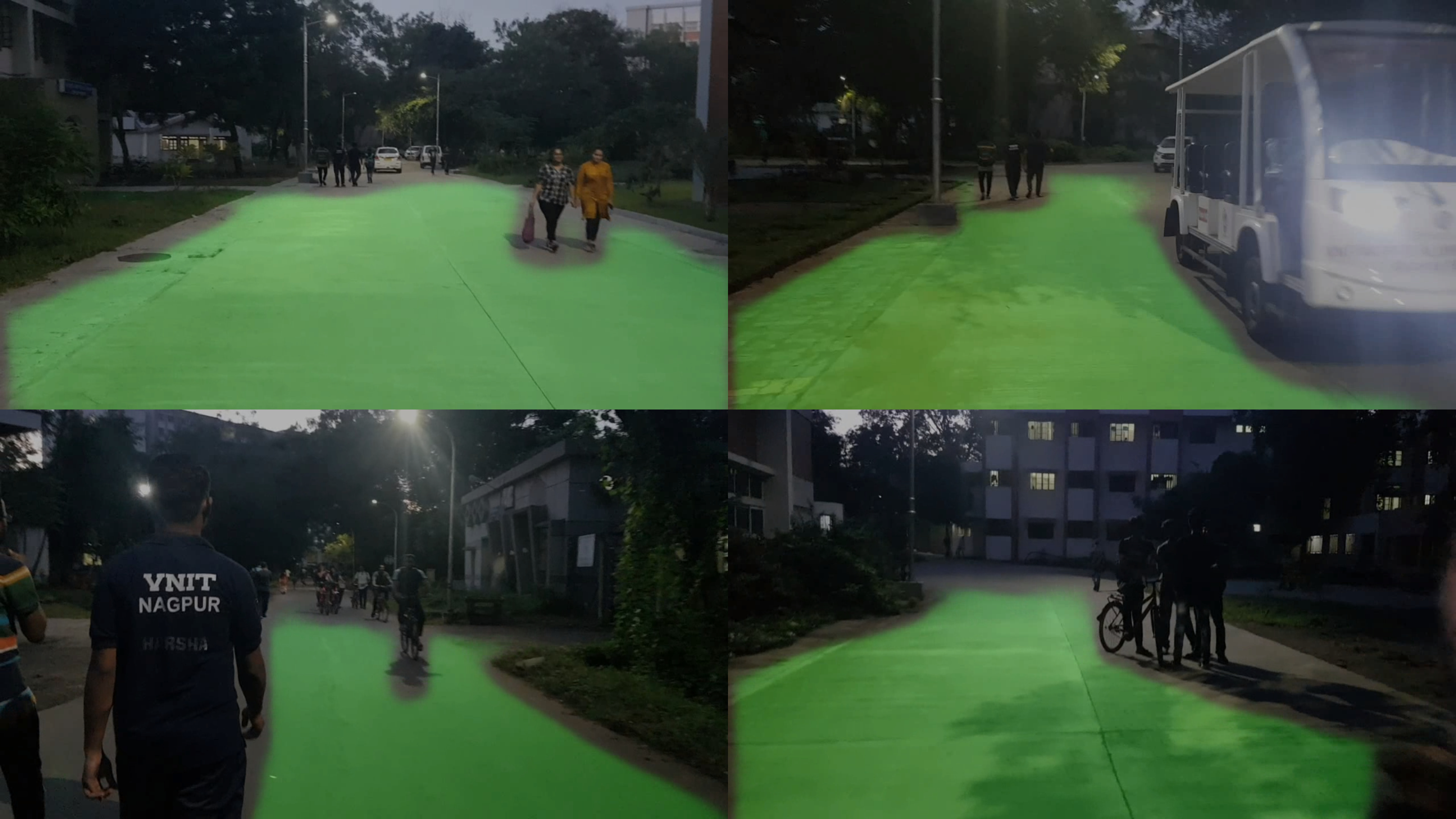}
\centering
\caption[ENet Results]{Results of ENet trained with Cityscapes dataset on VNIT campus images with green overlay representing the detected road pixels.}
\label{fig:enet_output}
\end{figure}

The result of these design choices is the reduction in memory and computational requirements while ensuring good performance. The results on some images from the VNIT campus are shown in Figure \ref{fig:enet_output}. Note that these images are captured using a handheld smartphone camera.

\subsection{Hardware Implementation}
Since this semantic segmentation model is to be deployed on the robot, it is essential to use a small, portable computer. Further, it should have enough computational power to process the images in real-time while also running the other algorithms (global and local path planning, localization, etc.) in parallel. Several such low-cost single-board computers are available in the market like the Raspberry Pi, ODroid, etc. However, they all lack a graphic processing unit (GPU) which severely limits the performance of the model while additionally introducing latency in the entire system. 

Thus, the choice of the onboard computer is limited to devices with a dedicated GPU. Nvidia has 2 products namely, Jetson TX1 and TX2 which have a dedicated GPU while maintaining a portable form factor. Thus, the Nvidia Jetson TX1 (since it's the cheaper option) is used as the onboard computer. The important hardware specifications of the Nvidia Jetson are as follows:
\begin{itemize}
    \item Quad-Core ARM Cortex A57 CPU
    \item 256-Core Nvidia Maxwell GPU
    \item 4 GB LPDDR4 memory (RAM and GPU VRAM is shared)
    \item 16 GB eMMC storage (with SD Card expansion slot)
    \item < 10 W power requirement
\end{itemize}

This small yet powerful device has its disadvantages. Most libraries, frameworks and software support for Intel x86 CPUs is very good. However, that is not true for ARM processors, mostly because they are not as widely used. Thus, the framework choices became limited, and the model was implemented in PyTorch because support for the TX1 was available readily in older versions of PyTorch.

\chapter{360° Vision using Catadioptric Camera Setup}
\label{chapter:catadioptric}

\section{Overview}
\subsection{Need of 360° Vision in Autonomous Vehicle.}
\par Building reliable vision capabilities for self-driving cars has been a major development hurdle. By combining a variety of sensors, however, developers have been able to create a detection system that can recognize a vehicle’s environment even better than human eyesight.

From photos to videos, cameras are the most accurate way to create a visual representation of the world, especially when it comes to autonomous vehicles.

Autonomous vehicles have to rely on cameras placed on every side - front, rear, left and right to stitch together a 360-degree view of their environment. Some have a wide field of view as much as 120 degrees and a shorter range. Others focus on a more narrow view to provide long-range visuals.

\subsection{Catadioptric System}

Catadioptric systems are those which make use of both lenses and mirrors for image formation. This contrast's with catoptric systems which use only mirrors and dioptric systems which use only lenses.

Panoramic images can be created from conventional catadioptric cameras. Ideal omnidirectional catadioptric cameras can provide images covering the whole view space. 

\section{Literature Review}

\subsection{Cameras with a Single Lens}

"Fish-eye" lenses provide a wide angle of view and can directly be used for panoramic imaging. A panoramic imaging system using a fisheye lens was described by Hall et al. in \cite{Hall}. A different example of an imaging system using a wide-angle lens was presented in \cite{a} where the panoramic camera was used to find targets in the scene. Fleck \cite{b} and Base et al. \cite{c} studied imaging models of fisheye lenses suitable for panoramic imaging.  

\subsection{Cameras with Single Mirror}

In 1970, Charles \cite{d} designed a mirror system for a single-lens reflex camera.  Various approaches on how to get panoramic images using different types of mirrors were described by Hamit \cite{e}. Gregus \cite{f} proposed a special lens to get a cylindrical projection directly without any image transformation. Chahl and Srinivasaan \cite{g} designed a convex mirror to optimize the quality of imaging. they derived the family of surfaces which preserve the linear relationship between the angle of incidence of light onto the surface and the angle of reflection into the conventional cameras.

\begin{figure}[h]
\includegraphics[width=0.6\textwidth]{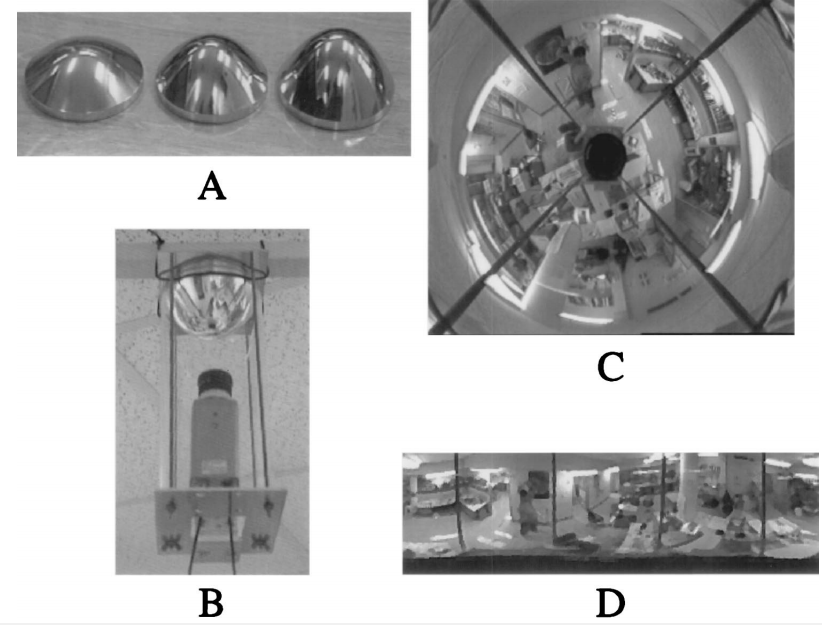}
\centering
\caption[Chahl et. al. Experimental Results]{Experimental implementation of the global imaging system. A, Three surfaces produced by turning aluminum on a CNC lathe. The mirrored finish was achieved by polishing with a metal polish of various grades. B, Assembled device enclosed in a glass tube for rigidity and to protect against dust. Internal reflection did not appear to be a major problem. C, Image produced by the
device. The field of view is approximately 240°. D, Image that resulted from unwarping the image, C. by Chahl et. al. in \cite{g}}
\label{fig:chahl_exp}
\end{figure}

\begin{figure}[t]
\includegraphics[width=0.7\textwidth]{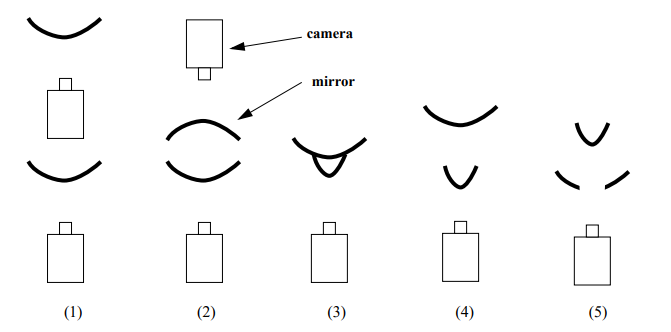}
\centering
\caption[Mark Ollis et. al. Catadioptric Configurations]{Side view of the five Catadioptric configurations examined by Mark Ollis et. al. \cite{Mark} in  (1) and (2) using two cameras. (3), (4) and (5) useing a single camera.}
\label{fig:Mark_exp}
\end{figure}

\begin{figure}[h]
\includegraphics[width=\textwidth]{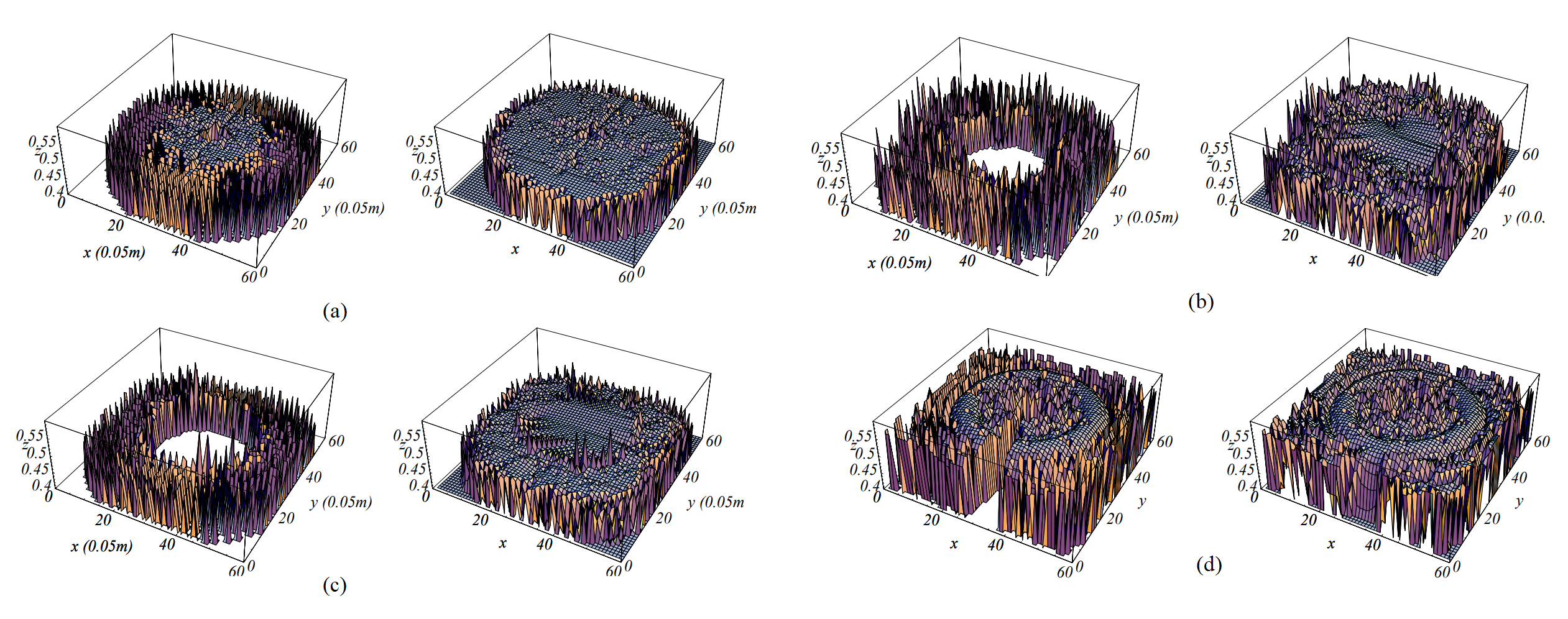}
\centering
\caption[Mark Ollis et. al. 3D Reconstruction Results]{3D view of reconstructed terrain using configuration 1. 3D points are integrated into a regularly spaced grid of height values.Raw terrain map \& Interpolated terrain map, using (a) configuration 1, (b) configuration 1, (c) configuration 1,(d) configuration 4 \& 5, by Mark Ollis et. al. \cite{Mark}}
\label{fig:Mark_res}
\end{figure}

\section{Proposed Omnivision (Catadioptric) Camera Setup}

\subsection{System Model}

\begin{figure}[h]
\includegraphics[width=0.75\textwidth]{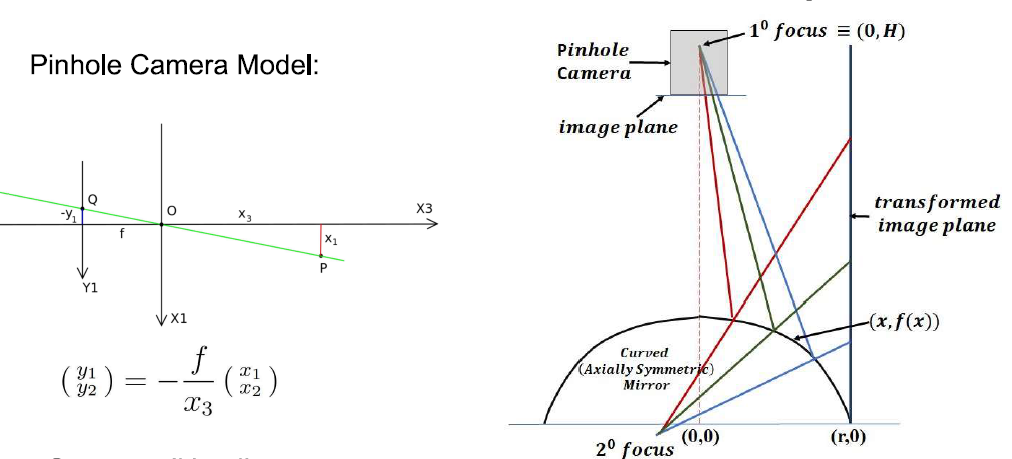}
\centering
\caption[Proposed Model]{Pinhole Camera Model and Proposed Omnivision (Catadioptric) Camera Setup}
\label{fig:Proposed_model}
\end{figure}

The proposed setup is given in Figure \ref{fig:Proposed_model}. The Pinhole camera model shown in the figure is used to focus and capture the reflected incoming light flux by an axially symmetric curved mirror at the bottom. The optical focus of the pinhole camera is defined as the primary $(1^{0})$ focus of the system and the converging point of the extended incoming light flux is estimated as secondary $(2^{0})$ focus of the system. 

\begin{figure}[h]
\includegraphics[width=\textwidth]{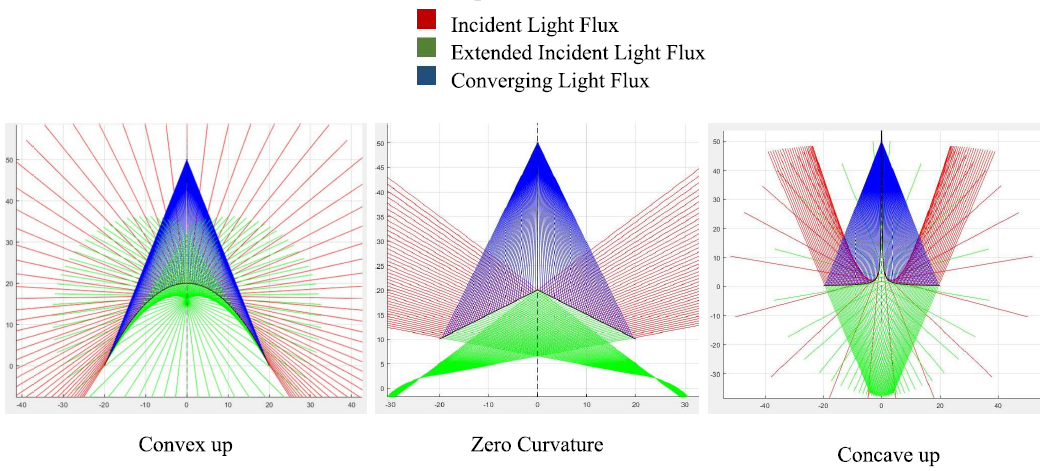}
\centering
\caption[Different Curved Mirror Topologies]{Light flux in different curved mirror typologies depending upon surface curvature.}
\label{fig:curvature}
\end{figure}

Figure \ref{fig:curvature} shows the performance of the proposed setup under the curved mirror typologies of three types of curvature conditions and the convention used for (1) Incoming Light Flux, (2) Extended Light Flux and (3) Converging Light Flux.

\subsection{General Cylindrical Projection of the Image}
\label{sec:general_cylindrical_projection}

Considering the reflected flux makes and image on the plane shown in Figure \ref{fig:Proposed_model} with respect to secondary focus, the curved mirror image to rectangular cylindrical projected image transformation equations is given by:

$$
f'(x) = -\tan{\theta} ,~~
\alpha-\theta = \Phi+\theta 
$$

$$
i.e.\phi = \alpha-2\theta ,
\tan{\alpha} = \frac{H-f(x)}{x} 
$$

$$
\tan{\phi} = \frac{A+B}{1-A.B},~~A=\frac{H-f(x)}{x},~~B=2.\frac{f'(x)}{1-f'(x)^2}
$$

\begin{equation}
h = f(x) + (r-x).{\frac{(1-f'(x)^2).(H-f(x))+2.x.f'(x)}{x.(1-f'(x)^2)-2.f'(x).(H-f(x))}}
\label{eqn:cylindrical_projection}
\end{equation}

\subsection{Suitable Shape for the curved Mirror}

Based on the equation \ref{eqn:cylindrical_projection}, the curved shape of the mirror is constrained due to certain conditions given by (a) The incoming flux lines should not intersect outside the mirror in order to have unique mapping on the rectangular frame, and (b) Extended flux lines must converge at the point inside the mirror to obtain the secondary focus of the system. 

Hence by considering the above conditions;

\begin{figure}%
    \centering
    {\includegraphics[width=0.4\textwidth]{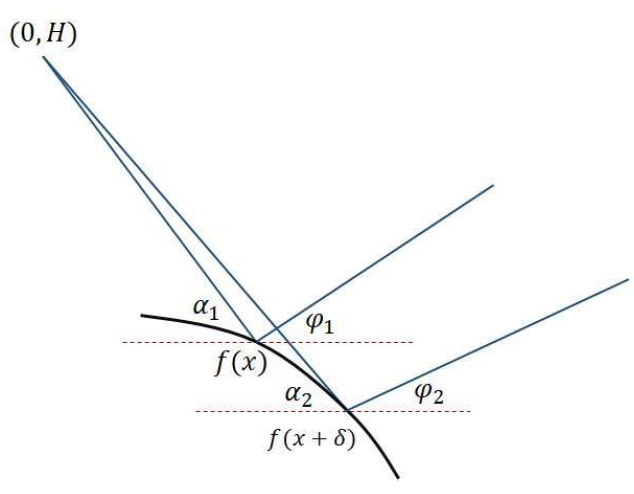}}%
    \qquad
    {\includegraphics[width=0.4\textwidth]{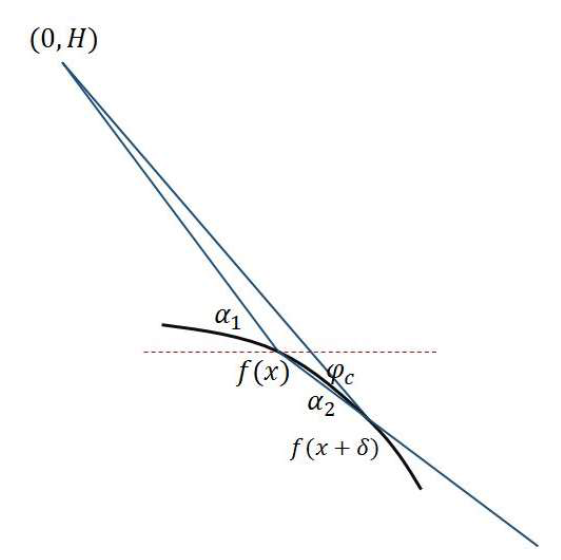}}%
    \caption[Mirror shape derivation.]{Suitable mirror shape derivation.}%
    \label{fig:derivation}%
\end{figure}

According to condition (a), 

$$
\text{for, }\alpha_1 > \alpha_2, ~~ \phi_1 > \phi_2 \text{ , i.e. } x_2 > x_1
$$

As shown in the Fig. \ref{fig:derivation}, let $\phi_c$ be the critical angle of reflection,

$$
\phi > \phi_c \\  
\therefore \tan{\phi} > \tan{\phi_c}, ~~ \phi_c , \phi \in (-\pi/2,\pi/2) 
$$

$$
\tan{\phi_c} = \frac{f(x)-f(x-\delta)}{\delta} 
$$

$$
\frac{f(x)-f(x-\delta)}{\delta} = \tan{\phi_{c1}} \leq \tan{\phi_1}
$$
$$
\frac{f(x+\delta)+f(x)}{\delta} = \tan{\phi_{c2}} \leq \tan{\phi_2} 
$$

$$
\text{but, } \tan{\phi_1} > \tan{\phi_2} 
$$

$$
\therefore \frac{f(x)-f(x-\delta)}{\delta} > \frac{f(x+\delta)+f(x)}{\delta} 
$$

for small $\delta$.

$$
{{f(x)} > {{\frac{f(x+{\delta})+f(x-{\delta)}}{2}}}} \\
{f'(x_1) > f'(x_2)} \quad \forall \; {x_2 > x_1} 
$$

if the function $f$ is monotonically decreasing,

$$
  \quad {f'(x_2) \leq f'(x_1)}\quad \forall  \; {x_2 > x_1} \\
  \quad {f'(x_1)\; ,f'(x_2) < 0}
$$

Hence, the curve function of the required shape of the mirror must be convex up.

\section{Analysis Over Curved Surface Topologies}

\begin{figure}[h]
\includegraphics[width=0.7\textwidth]{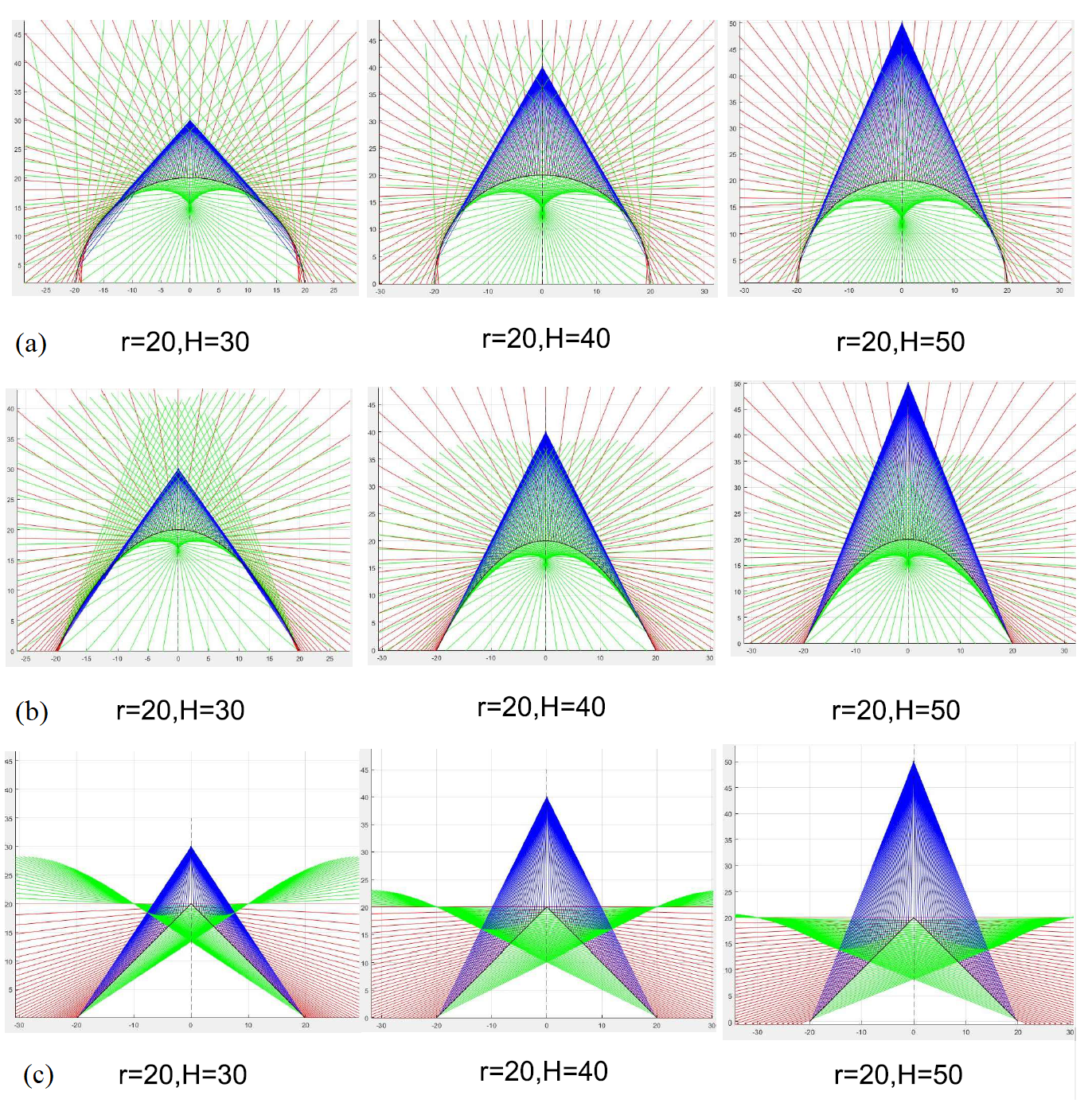}
\centering
\caption[Curved Mirrors]{Light flux plots of (a) Spherical Mirror, (b) Paraboloid Mirror, (c) Conical Mirror with parameters mentioned.}
\label{fig:curves}
\end{figure}

From the mathematical analysis discussed in the previous section, we developed the mathematical model for the system and compared the behavior of different geometrical shapes (sphere, cone, and paraboloid) under the set of parameters consisting of the height of the focus from the mirror center. The resultant plots obtained in MATLAB are given in the figure \ref{fig:curves}.

\begin{figure}[h]
\includegraphics[width=0.7\textwidth]{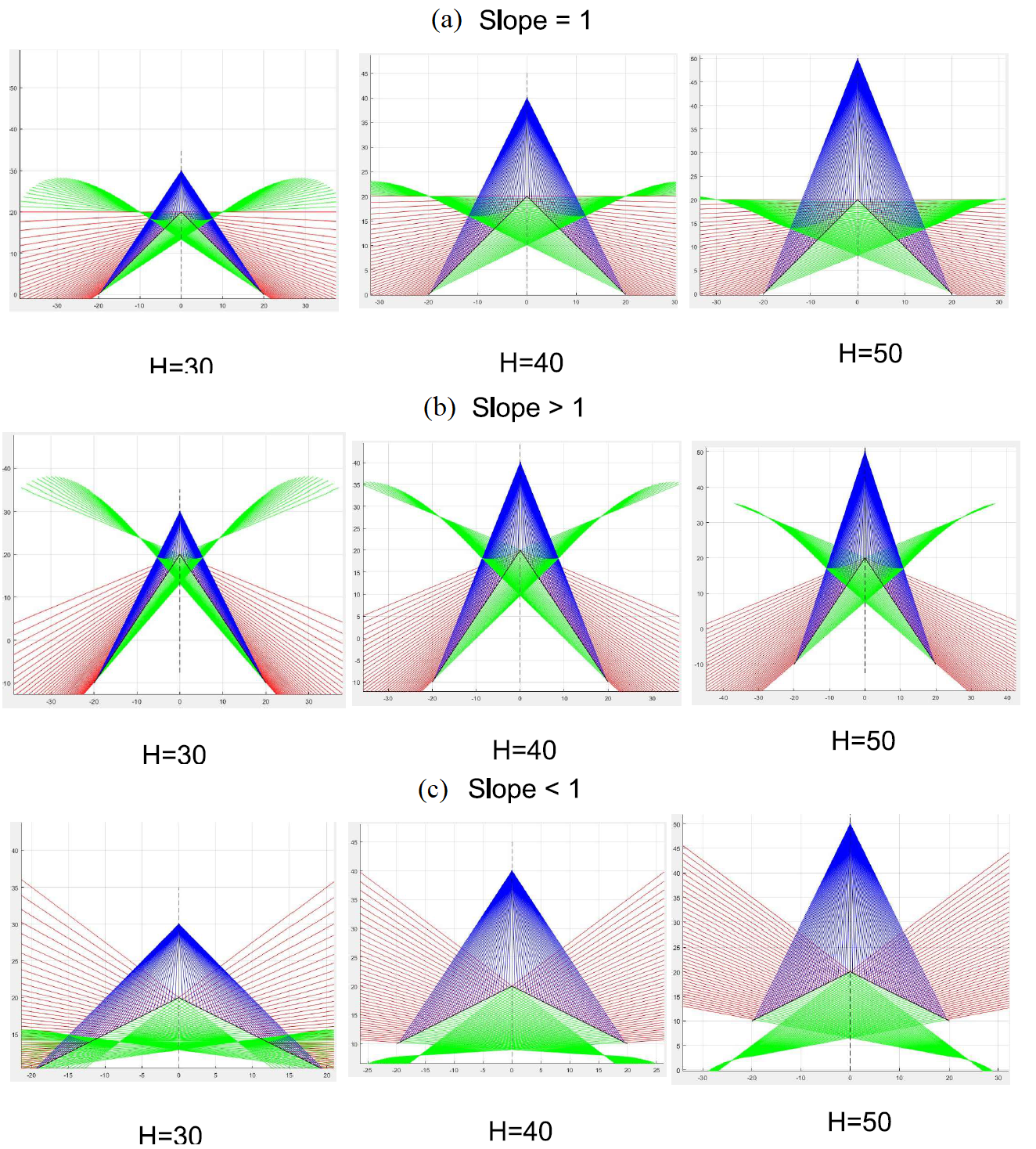}
\centering
\caption[Conical Mirrors]{Light flux plots of Conical mirrors with cross section slope (a) = 1, (b) > 1, (c) <1.}
\label{fig:cones}
\end{figure}

As shown in the figure \ref{fig:curves} the location secondary focus formed by Spherical and Paraboloid mirror is independent of the height H. But in the conical mirror secondary focus tends to go inside the mirror as height decreases. Also, the field of view in the conical mirror is ranged in the lower part parallel to the horizontal whereas in spherical and paraboloid it is located perpendicular to the horizontal reference. 

Similarly, the conical mirrors with different slopes of the cross-section and height 'H' are studied.

\section{Simulation Results}

Based on an analysis of the different curved mirror surfaces and conical mirrors, we developed CAD design of the conical mirror with suitable dimensions and other objects in surrounding using SolidWorks software. The figure shows the experimental setup and the rendered images taken from it. 

\begin{figure}[h]
\includegraphics[width=0.70\textwidth]{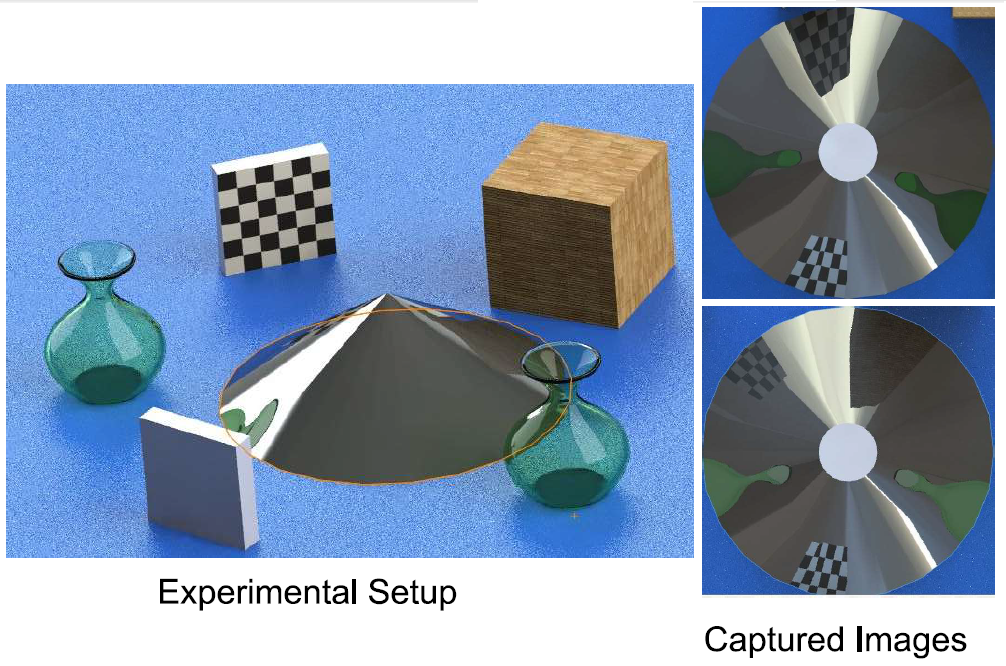}
\centering
\caption[Simulation Setup]{Experimental Setup and Captured Images from Conical Mirror.}
\label{fig:setup}
\end{figure}

\begin{figure}[h]
\includegraphics[width=0.75\textwidth]{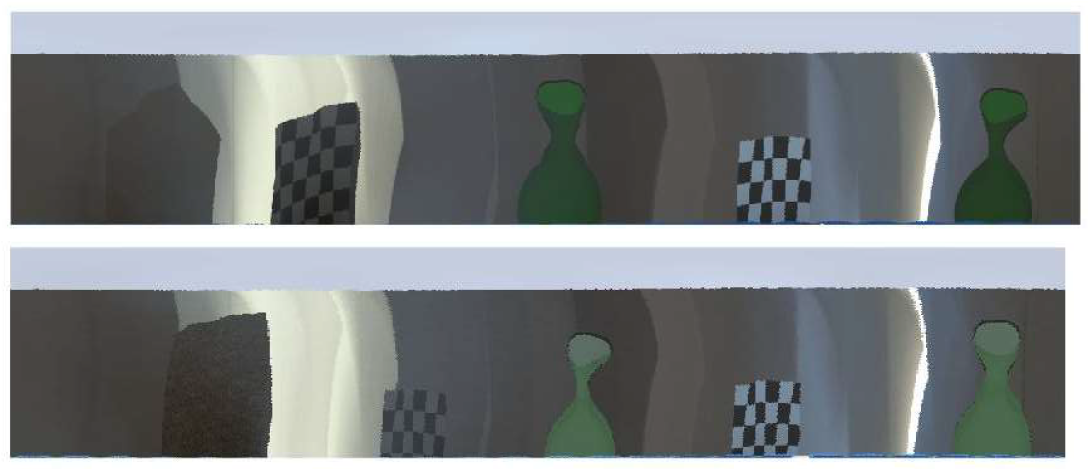}
\centering
\caption[Simulation Results]{Cylindrical Projection of images given in Figure \ref{fig:setup}}
\label{fig:result}
\end{figure}

Further, Cylindrical Projection is taken using equations mentioned in Subsection \ref{sec:general_cylindrical_projection}. The final results are shown in Figure \ref{fig:result}. 

\section{Conclusion}

\par The advantages of panoramic imaging with catadioptric setup are:
\begin{enumerate}
    \item Increased area coverage with single (or two) cameras.
    \item Simultaneous imaging of multiple targets.
    \item Instantaneous full-horizon detection.
    \item easier integration of various applications required for Autonomous Vehicles.
\end{enumerate}

   The proposed catadioptric system is economic over the present technologies providing the same information. The simplistic and static approach solves the problem more optimistically and economically. This idea has been discontinued for practical hardware implementation in this thesis since the manufacture of the conical reflector is too difficult for us to get a reliable reflector without distortions on the surface. Whereas, it will be very expensive if outsourced (or must be manufactured in bulk).

\chapter{Camera Lidar System Calibration}
\label{chapter:camera_lidar}
\section{Overview}
\subsection{Camera Calibration}
Camera Calibration also known as Camera Resectioning is the process of estimating internal parameters (also known as intrinsics) and external parameters (also known as extrinsics) of a camera. Internal parameter comprises of focal length, principal point, skew (if present), aspect ratio and lens distortions. External parameters comprises of 6 degrees of freedom associated with the camera in any arbitary coordinate system. In order to derive 3D (metric) information from a camera, calibration is an essential step. It allows photogrammetric measurements from images, distortion correction, 3D reconstruction, etc.

\subsection{LIDAR Calibration}
LIDAR stands for LIght Detection and Ranging. It is a device which provides accurate range measurements. It could be used in tasks such as localization, odometry, etc. In this project we are using a 2D LIDAR named YDLIDAR. It generates a planar point cloud of the scene. In order to fuse the information of LIDAR point cloud with the camera images, it is essential to register the devices together. By register, we mean to compute the transformation matrix between the Camera and a LIDAR.

\section{Pin Hole Camera Model}
\begin{figure}[h]
\includegraphics[width=\textwidth]{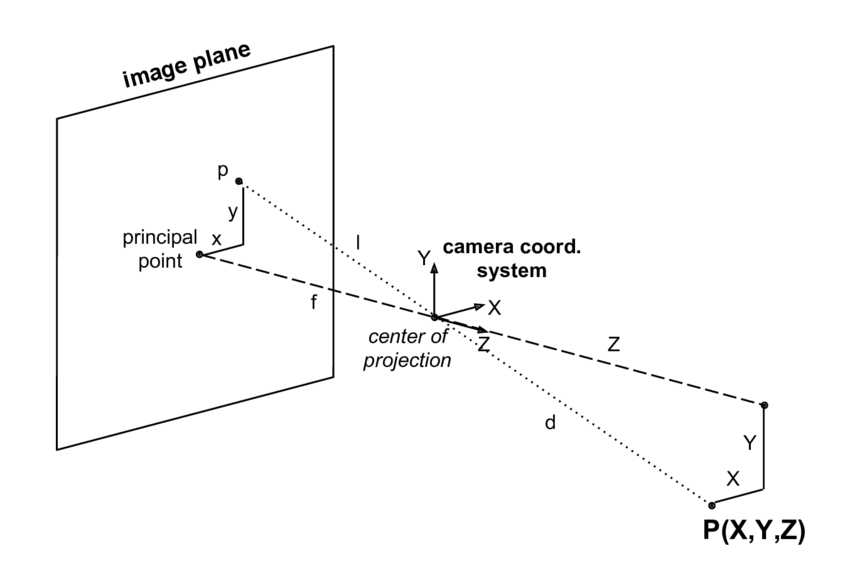}
\centering
\caption[Pinhole Camera]{Pinhole camera}
\label{fig:pinhole}
\end{figure}

In order to mathematically model digital cameras used in this project, it is important to gain understanding of a simple pinhole camera. A pinhole camera, unlike other cameras, do not have any lenses or mirrors, but just an aperture for light rays to enter and a film to capture the light. It is shown in Fig. \ref{fig:pinhole}. 

The image obtained from pin-hole camera is inverted and is projection of 3D world object onto a camera film. The size of the projected image is dependent on the distance of object from pin-hole as well as the distance of pin-hole from the camera film (also referred as focal length).

The relationship between the 3D object and its 2D image for a pin-hole camera from Fig. \ref{fig:pinhole} can be expressed as :

\begin{equation}
\frac{y}{-f} = \frac{Y}{Z}  \ \textrm{or} \ y = -f * \frac{Y}{Z}
\end{equation}

Similarly,
\begin{equation}
\frac{x}{-f} = \frac{X}{Z}  \ \textrm{or} \ x = -f * \frac{X}{Z}
\end{equation}

Negative sign in the equations indicates that the images are inverted.
\subsection{Extending pin-hole camera model to digital cameras}
The pin-hole model can be extended to cameras by making few changes : - 
\begin{itemize}
    \item The origin of image lies on the top left corner, so we must shift the projected coordinates with $(c_x, c_y)$, which represents the projection of camera center on the image plane plane
    \item In order to keep the equations 2.2.1 and 2.2.2 positive, we have assumed a virtual screen in front of the camera, thus eliminating the negative signs.
    \item The image recorded is not continuous but discretized by the image sensor into numerous pixels. Each pixel should ideally be square. However, it deviates from a square, in terms of aspect ratio and skew.
    \item After accounting all the parameters, the new transformation from camera coordinate system to image in homogeneous coordinates can be written as follows :
    \begin{equation}
    \lambda
    \begin{bmatrix} u\\
    v\\
    1
    \end{bmatrix} = 
     \begin{bmatrix}
     f_x & s & c_x \\
     0 & f_y & c_y \\
     0 & 0 & 1
     \end{bmatrix}
     \begin{bmatrix}
     X\\
     Y\\
     Z\\
     \end{bmatrix}
    \end{equation}
    In the above equation 2.2.3, $\lambda$ is equal to Z, which is the distance of the world point from the principal point.
    \item The $X, Y, Z$ in the equation 2.2.3, is expressed in camera coordinate system, in order to generalize it to any arbitrary Cartesian system we may introduce a Transformation Matrix $T$, from arbitrary coordinate system to camera coordinate system, thereby modifying equation 2.2.3 as :
    \begin{equation}
    \lambda
    \begin{bmatrix} u\\
    v\\
    1
    \end{bmatrix} = 
     \begin{bmatrix}
     f_x & s & c_x \\
     0 & f_y & c_y \\
     0 & 0 & 1
     \end{bmatrix}
     \begin{bmatrix}
     r_{11} & r_{12} & r_{13} & t_x \\
     r_{21} & r_{22} & r_{23} & t_y \\
     r_{31} & r_{32} & r_{33} & t_z
     \end{bmatrix}
     \begin{bmatrix}
     X\\
     Y\\
     Z\\
     \end{bmatrix}
    \end{equation}
    The first three columns of transformation matrix constitutes a rotation matrix and the last column is the translation vector.
    \item The pinhole model deviates as we move farther away from the image center, this effect can be accounted using polynomial model of distortion. We can also incorporate tangential distortions if necessary. 
    \end{itemize}

\section{Zhang's Method of Camera Calibration}
In order to calibrate our camera, we have used Zhang's method of camera calibration. This method requires us to capture images of an asymmetric checker board in different orientations and positions. The output of this method is a set of camera parameters:
\begin{itemize}
    \item The intrinsic parameters like:
    \begin{itemize}
        \item principal point ($c_x$, $c_y$)
        \item focal length in pixels ($f_x$, $f_y$)
        \item skew $s$ (if present)
        \item distortion coefficients
    \end{itemize}
    \item The extrinsic parameters like:
    \begin{itemize}
        \item The orientations of checkerboard w.r.t camera-coordinate system in different images
        \item The positions of checkerboard w.r.t camera-coordinate system in different images
    \end{itemize}
\end{itemize}

This information will be used in further processing.

\section{Camera LIDAR Extrinsic Calibration}
In order to extrinsically calibrate LIDAR, we have implemented the algorithm presented in ??. This method requires us to capture images as well as the point clouds of the scene in which checker-board is placed. The orientation of checkerboard is changed after every capture.
Once the capture is completed, we manually segment the points from every point-cloud of LIDAR that belongs to checker-board. Then we calibrate the camera from the images which were captured using Zhang's method described in Section 2.3. 
For every pose of checkerboard we compute the normal N in camera cordinate system using the equation:
\begin{equation}
    N = -R_3(R_3^T.t)
\end{equation}
Here, $R_3$ represents the third column of rotation matrix from equation 2.2.4 and t represents the translation vector.
The equation of the checkerboard plane can then be written as 
\begin{equation}
    N.x = \|N\|^2
\end{equation}
Here, $x$ denotes the points lying on the plane, in camera system.
Now lets assume that the transformation between the points in camera and lidar is represented as : -
\begin{equation}
P_l = \phi P_c + \delta   \ \textrm{or} \ P_c = \phi^{-1}(P_l - \delta)  
\end{equation} 
Here, $P_l$ represents the point in Lidar system and $P_c$ in Camera System. $\phi$ is the rotation matrix and $\delta$ is translation vector between them. 
Substituting equation 2.4.3 in 2.4.2, we get
\begin{equation}
  N.\phi^{-1}(P_l - \delta) = \|N\|^2  
\end{equation}
Equation 2.4.4 can further be simplified as :
\begin{equation}
N.HP_l = \|N\|^2 
\end{equation}
Where, 
\begin{equation}
    H = \phi^{-1}\begin{bmatrix} 1 & 0 & -\delta_x \\
 1 & 0 & -\delta_y \\
  1 & 0 & -\delta_z
\end{bmatrix}
\end{equation}
We can solve for $H$ with every pose with multiple LIDAR points linearly using least squares. 

After determining $H = [H_1, H_2, H_3]$, $\phi$ and $\delta$ can be computed as
 \begin{equation}
     \phi = [H_1, -H_1 \times H_2, H_2]^T
 \end{equation}
 \begin{equation}
     \delta = -\phi.H_3
 \end{equation}
The solution obtained can then be refined by non-linear optimization using Levenberg-Marquardt algorithm.
\chapter{Conclusion and Future Work}

\section{Conclusion}

In this thesis, the construction of the prototype and the tasks of perception, mapping, localization and planning for the Autonomous Delivery Robot are demonstrated. We have implemented and tested the various modules required in an autonomous robot operating in open environments. The issues faced in design and operation of the robot in outdoor environments have been discussed in detail in this thesis. We aimed to create a completely autonomous robotic system, robust enough to operate smoothly on roads with proper localization and planning. This would solve the problem of robots being manually controlled by a human operator and help in carrying out tasks for humans to make their lives easier.

Although the purpose of the robot in this thesis is aligned towards a delivery robot, the pipeline explained in the thesis can be used for any robots operating autonomously in an outdoor environment. For example, the presented system can also play a significant role in medical assistance applications such as in the current COVID-19 pandemic situation, autonomous food and medicine delivery as well as sample collection in critical areas.

\section{Future Work}
This thesis demonstrates an end-to-end pipeline for an autonomous robot operating in an outdoor environment. The current prototype is centered on the software portion and theory implementation rather than the actual functionality of the delivery vehicle. Therefore, in the future, the hardware will be more focused on delivery vehicle aspects such as effective human-machine interface as well as the dashboard,  distant monitoring system and sound fail-safe management, long run time, and the larger payload capacity. However, highly complex situations were avoided while testing. The outdoor environment is highly unpredictable due to dynamic obstacles such as pedestrians, animals, vehicles, etc. Further research could be carried out to develop highly optimized and robust planning algorithms to overcome the mentioned issues in an efficient manner. The robot could be made more robust to operate in different weather conditions like rain, snow, etc. and on uneven road terrains. The future work can include developing efficient algorithms or choose appropriate sensors to tackle the issues faced in localization of the robot in outdoor environment as explained in Chapter \ref{chapter:localization}. Instead of using a single channel Lidar for localization, a multi-channel Lidar could be used to provide a well defined 3D point cloud. The map can be refined using additional layers as mentioned in the literature review to construct a precise 3D map of the environment containing very useful information. The autonomous robot and the underlying algorithms explained in this thesis can be used as a base platform to conduct further research in each independent module.
\chapter*{Acknowledgement}
\par First and foremost, we express our sincere gratitude to our guide, Dr. Ashwin Dhabale, Assistant Professor, Department of Electrical Engineering, for his continuous support and guidance throughout the course of the project. 


We would also like to thank Dr. Shital Chiddarwar of Mechanical Engineering Department and the Professor-in-charge of IvLabs - the Robotics Club of VNIT for providing space to work in IvLabs and for GPU access for training our models. 






\printbibliography

\end{document}